\documentclass[10pt,twocolumn,letterpaper]{article}

\usepackage{iccv}
\usepackage{times}
\usepackage{epsfig}
\usepackage{graphicx}
\usepackage{amsmath}
\usepackage{amssymb}

\usepackage{animate}
\usepackage{subfig}
\usepackage{array}
\usepackage{multirow}
\usepackage{mathabx}
\usepackage{microtype}
\usepackage{enumitem}
\usepackage{xspace}
\usepackage{xcolor}
\usepackage{color,soul}
\usepackage{multicol}
\usepackage{colortbl}
\usepackage{tabularx,booktabs}
\usepackage{cancel}
\usepackage{booktabs}
\usepackage{float}
\usepackage{pifont}
\usepackage{adjustbox}
\usepackage{epigraph}

\usepackage{caption}
\newcommand{\sota}{state-of-the-art\xspace}

\newcommand{\bp}{\mathbf{p}} %
\newcommand{\bv}{\mathbf{v}} %
\newcommand{\bc}{\mathbf{c}} %
\newcommand{\bz}{\mathbf{z}} %
\newcommand{\bo}{\mathbf{o}} %
\newcommand{\br}{\mathbf{r}} %

\let\oldFootnote\footnote
\newcommand\nextToken\relax
\renewcommand\footnote[1]{%
    \oldFootnote{#1}\futurelet\nextToken\isFootnote}
\newcommand\isFootnote{%
    \ifx\footnote\nextToken\textsuperscript{,}\fi}

\definecolor{overview_orange}{HTML}{FFA726}
\definecolor{overview_green}{HTML}{9CCC65}
\definecolor{overview_purple}{HTML}{D877E9}
\colorlet{overview_red}{red!35}
\definecolor{overview_gray}{HTML}{BEBEBE}

\newcommand{\acrobat}{}

\newcolumntype{Y}{>{\centering\arraybackslash}X}

\newcommand{\PreserveBackslash}[1]{\let\temp=\\#1\let\\=\temp}
\newcolumntype{C}[1]{>{\PreserveBackslash\centering}p{#1}}
\newcolumntype{R}[1]{>{\PreserveBackslash\raggedleft}p{#1}}
\newcolumntype{L}[1]{>{\PreserveBackslash\raggedright}p{#1}}

\newcolumntype{F}[1]{%
 >{\vbox to 5ex\bgroup\vfill\centering}%
 p{#1}%
 <{\egroup}}

\makeatletter
\let\UrlSpecialsOld\UrlSpecials
\def\UrlSpecials{\UrlSpecialsOld\do\/{\Url@slash}\do\_{\Url@underscore}}%
\def\Url@slash{\@ifnextchar/{\kern-.11em\mathchar47\kern-.2em}%
    {\kern-.0em\mathchar47\kern-.08em\penalty\UrlBigBreakPenalty}}
\def\Url@underscore{\nfss@text{\leavevmode \kern.06em\vbox{\hrule\@width.3em}}}
\makeatother

\usepackage[pagebackref=true,breaklinks=true,letterpaper=true,colorlinks,bookmarks=false]{hyperref}

\iccvfinalcopy %

\ificcvfinal\fi

\begin{document}

\title{GANcraft: Unsupervised 3D Neural Rendering of Minecraft Worlds}

\author{Zekun Hao$^{\ast\dagger}$, Arun Mallya$^\ast$, Serge Belongie$^\dagger$, Ming-Yu Liu$^\ast$ \\
$^\ast$NVIDIA, $^\dagger$Cornell University \\
{\small\tt \{hz472, sjb344\}@cornell.edu, \{amallya, mingyul\}@nvidia.com}
}

\twocolumn[{%
\renewcommand\twocolumn[1][]{#1}%
\maketitle
\begin{center}
    
\vspace{-15pt}
\centering

\begin{adjustbox}{max width=\textwidth}
\setlength{\tabcolsep}{0pt}
\begin{tabular}{cc}
    \href{https://nvlabs.github.io/GANcraft/videos/0366_3.mp4}{\includegraphics{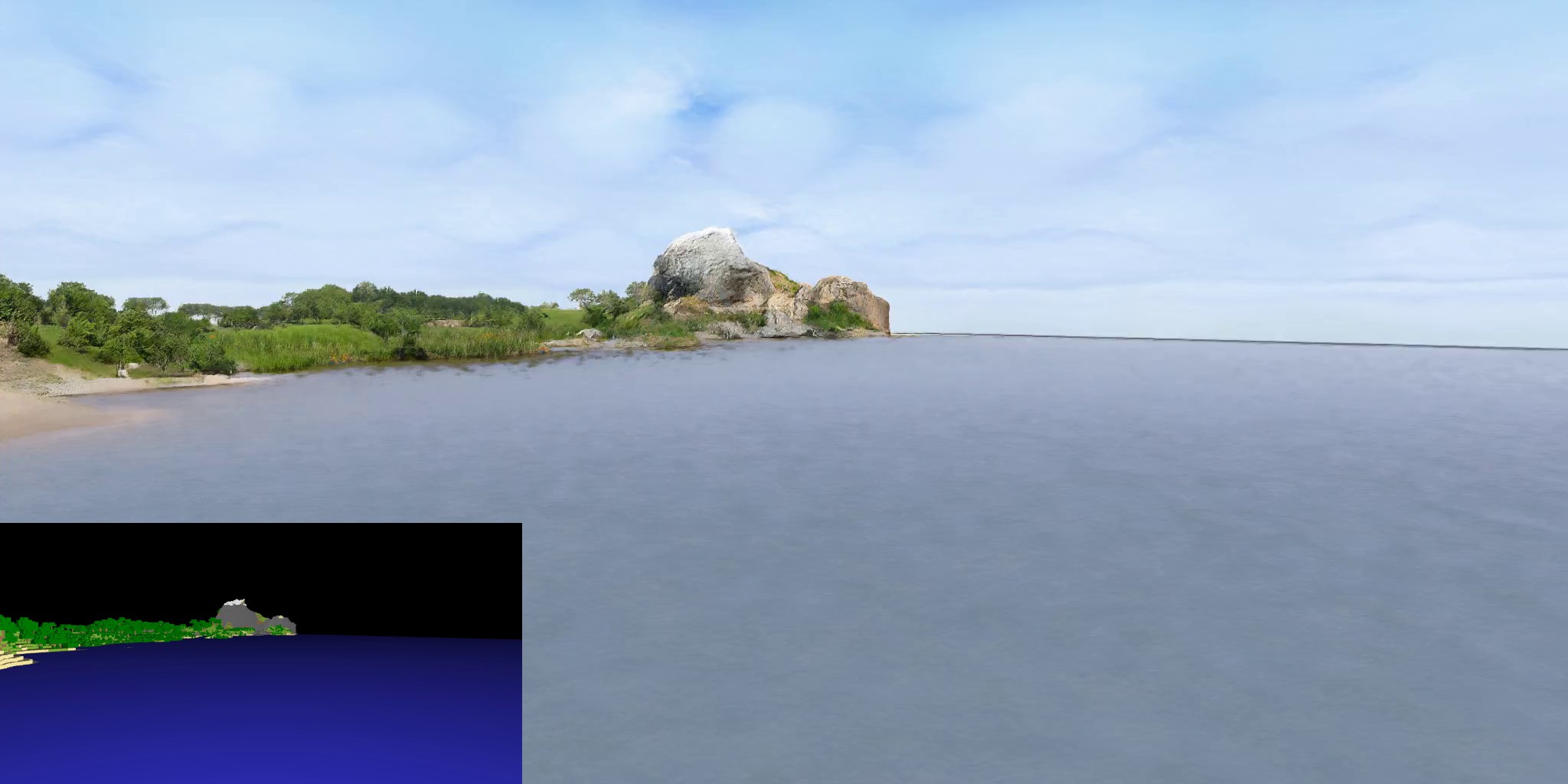}} &

    \href{https://nvlabs.github.io/GANcraft/videos/3117_2.mp4}{\includegraphics{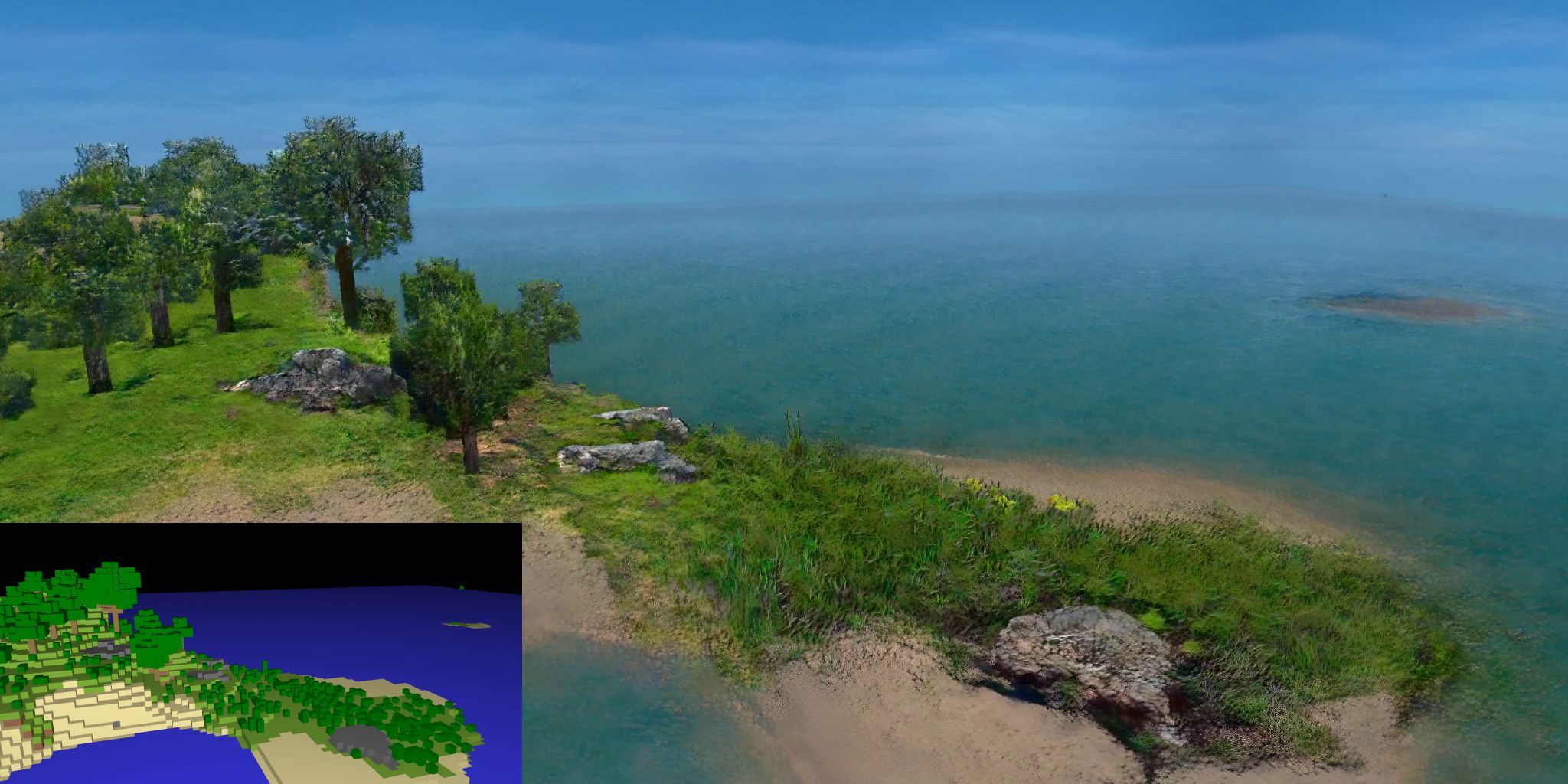}}

\end{tabular}
\end{adjustbox}
\captionof{figure}{Given a semantically-labeled block world as input (insets), GANCraft generates high-resolution view-consistent realistic outputs. It unsupervisedly learns to translate the input world to a realistic-looking world in the absence of paired training data across these two worlds. \emph{Click on image to play video in web browser.} \acrobat}
\vspace{10pt}
\label{fig:teaser}

\end{center}%
}]

\ificcvfinal\thispagestyle{empty}\fi

\begin{abstract}
   We present GANcraft, an unsupervised neural rendering framework for generating photorealistic images of large 3D block worlds such as those created in Minecraft. Our method takes a semantic block world as input, where each block is assigned a semantic label such as dirt, grass, or water. We represent the world as a continuous volumetric function and train our model to render view-consistent photorealistic images for a user-controlled camera. In the absence of paired ground truth real images for the block world, we devise a training technique based on pseudo-ground truth and adversarial training. This stands in contrast to prior work on neural rendering for view synthesis, which requires ground truth images to estimate scene geometry and view-dependent appearance. In addition to camera trajectory, GANcraft allows user control over both scene semantics and output style. Experimental results with comparison to strong baselines show the effectiveness of GANcraft on this novel task of photorealistic 3D block world synthesis. The project website is available at \href{https://nvlabs.github.io/GANcraft/}{\texttt{https://nvlabs.github.io/GANcraft/}}.
   \vspace{-24pt}
\end{abstract}

\section{Introduction}
\label{sec:introduction}

\noindent\hfill\emph{Imagine a world where every Minecrafter is a 3D painter!}

Advances in 2D image-to-image translation~\cite{gaugan,isola2017image,park2019semantic} have enabled users to \emph{paint} photorealistic images by drawing simple sketches similar to those created in Microsoft Paint.
Despite these innovations, creating a realistic 3D scene remains a painstaking task, out of the reach of most people.
It requires years of expertise, professional software, a library of digital assets, and a lot of development time.
In contrast, building 3D worlds with blocks, say physical LEGOs or their digital counterpart, is so easy and intuitive that even a toddler can do it.
Wouldn't it be great if we could build a simple 3D world made of blocks representing various materials
(like Fig.~\ref{fig:teaser}\,(insets)),
feed it to an algorithm, and receive a realistic looking 3D world featuring tall green trees, ice-capped mountains, and the blue sea
(like Fig.~\ref{fig:teaser})?
With such a method, we could perform \emph{world-to-world} translation to convert the worlds of our imagination to reality.
Needless to say, such an ability would have many applications, from entertainment and education, to rapid prototyping for artists.

\begin{table}[t]
    \setlength{\tabcolsep}{2.5pt}
    \centering
    \begin{adjustbox}{max width=\columnwidth}
    \begin{tabular}{cccc}
        \toprule
        Representative & 3D view & Across & W/o paired \\
        method & consistent? & worlds? & data? \\
        \midrule
        CycleGAN~\cite{zhu2017unpaired}, MUNIT~\cite{huang2018multimodal} & \ding{55} & \ding{51} & \ding{51}\\
        pix2pix~\cite{isola2017image}, SPADE~\cite{park2019semantic} & \ding{55} & \ding{51} & \ding{55} \\
        wc-vid2vid~\cite{mallya2020world} & \ding{51} & \ding{51} & \ding{55} \\
        NeRF~\cite{mildenhall2020nerf}, NSVF~\cite{liu2020neuralsparse} & \ding{51} & \ding{55} & \ding{55} \\
        GANcraft (ours) & \ding{51} & \ding{51} & \ding{51}\\
        \bottomrule
    \end{tabular}
    \end{adjustbox}
    
    \caption{Given a Minecraft world, our goal is to train a neural renderer that can convert any camera trajectory in the Minecraft world to a sequence of view-consistent images in the real world, at test time. The training needs to be achieved without paired Minecraft--real data as it does not exist. Among prior work, only unsupervised image-to-image translation methods such as CycleGAN~\cite{zhu2017unpaired} and MUNIT~\cite{huang2018multimodal} can work in this setting. However, they do not generate 3D view-consistent outputs. Neural radiance field-based methods like NeRF~\cite{mildenhall2020nerf} and NSVF~\cite{liu2020neuralsparse} are suited for novel view synthesis. They cannot handle the Minecraft--real domain gap. All other prior works require paired training data unavailable in our setting. GANcraft, our proposed method, can generate 3D view-consistent Minecraft-to-real synthesis results without paired Minecraft--real training data.
    }
    \label{table:overview_comparison}
\end{table}

In this paper, we propose GANcraft, a method that produces realistic renderings of semantically-labeled 3D block worlds, such as those from Minecraft (\url{www.minecraft.net}).
Minecraft, the best-selling video game of all time with over 200 million copies sold and over 120 million monthly users~\cite{minecraft_user_stats}, is a sandbox video game in which a user can explore a procedurally-generated 3D world made up of blocks arranged on a regular grid, while modifying and building structures with blocks.
Minecraft provides blocks representing various building materials---grass, dirt, water, sand, snow, \etc. Each block is assigned a simple texture,
and the game is known for its distinctive cartoonish look.
While one might discount Minecraft as a simple game with simple mechanics, Minecraft is, in fact, a very popular 3D content creation tool.
Minecrafters have faithfully recreated \href{https://www.planetminecraft.com/project/greenfield---new-life-size-city-project/}{large cities} and famous landmarks including the \href{https://www.grabcraft.com/minecraft/eiffel-tower-paris/towers}{Eiffel Tower}!
The block world representations are intuitive to manipulate and this makes it well-suited as the medium for our world-to-world translation task.
We focus on generating natural landscapes, which was also studied in several prior work in image-to-image translation~\cite{gaugan,park2019semantic}.

At first glance, generating a 3D photorealistic world from a semantic block world seems to be a task of translating a sequence of projected 2D segmentation maps of the 3D block world, and is a direct application of image-to-image translation . This approach, however, immediately runs into several serious issues. First, obtaining paired ground truth training data of the 3D block world, segmentation labels, and corresponding real images is extremely costly if not impossible. Second, existing image-to-image translation models~\cite{huang2018multimodal,park2019semantic,wang2018high,zhu2017unpaired} do not generate consistent views~\cite{mallya2020world}. Each image is translated independent of the others.

While the recent world-consistent vid2vid work~\cite{mallya2020world} overcomes the issue of view-consistency, it requires paired ground truth 3D training data. Even the most recent neural rendering approaches based on neural radiance fields
such as NeRF~\cite{mildenhall2020nerf}, NSVF~\cite{liu2020neuralsparse}, and NeRF-W~\cite{martin2020nerf}, require real images of a scene and associated camera parameters, and are best suited for view interpolation. As there is no paired 3D and ground truth real image data, as summarized in Table~\ref{table:overview_comparison}, none of the existing techniques can be used to solve this new task. This requires us to employ ad hoc adaptations to make our problem setting as similar to these methods' requirements as possible, \eg training them on real segmentations instead of Minecraft segmentations.

In the absence of ground truth data, we propose a framework to train our model using \emph{pseudo}-ground truth photorealistic images for sampled camera views.
Our framework uses ideas from image-to-image translation and improves upon work in 3D view synthesis to produce view-consistent photorealistic renderings of input Minecraft worlds as shown in
Fig.~\ref{fig:teaser}.
Although we demonstrate our results using Minecraft, our method works with other 3D block world representations, such as voxels. We chose Minecraft because it is a popular platform available to a wide audience.

\noindent Our key contributions include:
\begin{itemize}[noitemsep,topsep=0pt,parsep=0pt,partopsep=0pt]
    \item The novel task of producing view-consistent photorealistic renderings of user-created 3D semantic block worlds, or \emph{world-to-world} translation, a 3D extension of image-to-image translation.

    \item A framework for training neural renderers in the absence of ground truth data. This is enabled by using \emph{pseudo}-ground truth images generated by a pretrained image synthesis model (Section~\ref{subsec:pseudo_gt}).

    \item A new neural rendering network architecture trained with adversarial losses (Section~\ref{subsec:network_arch}), that extends recent work in 2D and 3D neural rendering~\cite{huang2017adain,liu2020neuralsparse,martin2020nerf,mildenhall2020nerf,niemeyer2020giraffe} to produce \sota results which can be conditioned on a style image (Section~\ref{sec:experiments}).
\end{itemize}

\section{Related Work}
\label{sec:related_work}

\noindent{\bf 2D image-to-image translation.} The GAN framework~\cite{goodfellow2014generative} has enabled multiple methods to successfully map an image from one domain to another with high fidelity, \eg, from input segmentation maps to photorealistic images. This task can be performed in the supervised setting~\cite{isola2017image,lee2020maskgan,liu2019learning,park2019semantic,wang2018high,zhu2017toward}, where
example pairs of corresponding images are available, as well as the unsupervised setting~\cite{choi2017stargan,huang2018multimodal,liu2016unsupervised,liu2019few,liu2019learning,schonfeld2021you,zhu2017unpaired}, where only two sets of images are available.
Methods operating in the supervised setting use stronger losses such as the L$_1$ or perceptual loss~\cite{johnson2016perceptual}, in conjunction with the adversarial loss. As paired data is unavailable in the unsupervised setting, works typically rely on a shared-latent space assumption~\cite{liu2016unsupervised} or cycle-consistency losses~\cite{zhu2017unpaired}.
For a comprehensive overview of image-to-image translation methods, please refer to the survey of Liu~\etal~\cite{liu2020generative}.

Our problem setting naturally falls into the unsupervised setting as we do not possess real-world images corresponding to the Minecraft 3D world. To facilitate learning a view-consistent mapping, we employ \emph{pseudo}-ground truths during training, which are predicted by a pretrained supervised image-to-image translation method.

\smallskip
\noindent{\bf Pseudo-ground truths} were first explored in prior work on self-training, or bootstrap learning~\cite{mcclosky2006effective,yarowsky1995unsupervised}\footnote{See {\scriptsize \url{https://ruder.io/semi-supervised/}} for an overview.}. More recently, this technique has been adopted in several unsupervised domain adaptation works~\cite{choi2019pseudo,lee2013pseudo,shu2018dirt,tang2012shifting,xie2018learning,zhang2018collaborative,zou2018unsupervised}. They use a deep learning model trained on the `source' domain to obtain predictions on the new `target' domain, treat these predictions as ground truth labels, or \emph{pseudo labels}, and finetune the deep learning model on such self-labeled data. 

In our problem setting, we have segmentation maps obtained from the Minecraft world but do not possess the corresponding real image. We use SPADE~\cite{park2019semantic}, a conditional GAN model, trained for generating landscape images from input segmentation maps to generate pseudo ground truth images. This yields the pseudo pair: input Minecraft segmentation mask and the corresponding pseudo ground truth image. 
The pseudo pairs enable us to use stronger supervision such as L$_1$, L$_2$, and perceptual~\cite{johnson2016perceptual} losses in our world-to-world translation framework, resulting in improved output image quality. 
This idea of using pretrained GAN models for generating training data has also been explored in the very recent works of Pan~\etal~\cite{pan2020gan2shape} and  Zhang~\etal~\cite{zhang2021image}, which use a pretrained StyleGAN~\cite{karras2018style,karras2020analyzing} as a multi-view data generator to train an inverse graphics model.

\smallskip
\noindent{\bf 3D neural rendering.} A number of works have explored combining the strengths of the traditional graphics pipeline, such as 3D-aware projection, with the synthesis capabilities of neural networks to produce view-consistent outputs. By introducing differentiable 3D projection and using trainable layers that operate in the 3D and 2D feature space, several recent methods~\cite{aliev2019neural,henzler2019escaping,nguyen2019hologan,nguyen2020blockgan,sitzmann2019deepvoxels,wiles2019synsin} are able to model the geometry and appearance of 3D scenes from 2D images. 
Some works have successfully combined neural rendering with adversarial training~\cite{henzler2019escaping,nguyen2019hologan,nguyen2020blockgan,niemeyer2020giraffe,schwarz2020graf}, thereby removing the constraint of training images having to be posed and from the same scene. However, the under-constrained nature of the problem limited the application of these methods to single objects, synthetic data, or small-scale simple scenes.
As shown later in Section~\ref{sec:experiments}, we find that adversarial training alone is not enough to produce good results in our setting. This is because our input scenes are larger and more complex, the available training data is highly diverse, and there are considerable gaps in the scene composition and camera pose distribution between the block world and the real images. 

Most recently, NeRF~\cite{mildenhall2020nerf} demonstrated state-of-the-art results in novel view synthesis by encoding the scene in the weights of a neural network that produces the volume density and view-dependent radiance at every spatial location. 
The remarkable synthesis ability of NeRF has inspired a large number of follow-up works which have tried to improve the output quality~\cite{liu2020neuralsparse,zhang2020nerf}, make it faster to train and evaluate~\cite{lindell2020autoint,liu2020neuralsparse,neff2021donerf,rebain2020derf,tancik2020learned}, extend it to deformable objects~\cite{du2020neural,li2020neural,park2020deformable,pumarola2020d,xian2020space}, account for lighting~\cite{boss2020nerd,bi2020neural,martin2020nerf,srinivasan2020nerv} and compositionality~\cite{guo2020object,niemeyer2020giraffe,ost2020neural,yuan2020star}, as well as add generative capabilities~\cite{chan2020pi,schwarz2020graf,niemeyer2020giraffe}.

Most relevant to our work are NSVF~\cite{liu2020neuralsparse}, NeRF-W~\cite{martin2020nerf}, 
and GIRAFFE~\cite{niemeyer2020giraffe}. NSVF~\cite{liu2020neuralsparse} reduces the computational cost of NeRF by representing the scene as a set of voxel-bounded implicit fields organized in a sparse voxel octree, which is obtained by pruning an initially dense cuboid made of voxels. 
NeRF-W~\cite{martin2020nerf} learns image-dependent appearance embeddings allowing it to learn from unstructured photo collections, and produce style-conditioned outputs. 
These works on novel view synthesis learn the geometry and appearance of scenes given ground truth posed images. 
In our setting, the problem is inverted --- we are given coarse voxel geometry and segmentation labels as input, without any corresponding real images. 

Similar to NSVF~\cite{liu2020neuralsparse}, we assign learnable features to each corner of the voxels to encode geometry and appearance. 
In contrast, we do not learn the 3D voxel structure of the scene from scratch, but instead implicitly refine the provided coarse input geometry (\eg shape and opacity of trees represented by blocky voxels) during the course of training. Prior work by Riegler~\etal~\cite{riegler2020free} also used a mesh obtained by multi-view stereo as a coarse input geometry.
Similar to NeRF-W~\cite{martin2020nerf}, we use a style-conditioned network. This allows us to learn consistent geometry while accounting for the view inconsistency of SPADE~\cite{park2019semantic}.
Like neural point-based graphics~\cite{aliev2019neural} and GIRAFFE~\cite{niemeyer2020giraffe}, we use differentiable projection to obtain features for image pixels, and then use a CNN to convert the 2D feature grid to an image.
Like GIRAFFE~\cite{niemeyer2020giraffe}, we use an adversarial loss in training. We, however, learn on large, complex scenes and produce higher-resolution outputs (1024$\times$2048 original image size in Fig.~\ref{fig:teaser}, v/s 64$\times$64 or 256$\times$256 pixels in GIRAFFE), in which case adversarial loss alone fails to produce good results.

\section{Neural Rendering of Minecraft Worlds}
\label{sec:method}

\begin{figure*}[t]
\centering
\setlength{\belowcaptionskip}{-4pt}

\begin{minipage}{.46\textwidth}
    \includegraphics[width=\textwidth, trim={0.2cm 11cm 13cm 3cm}, clip]{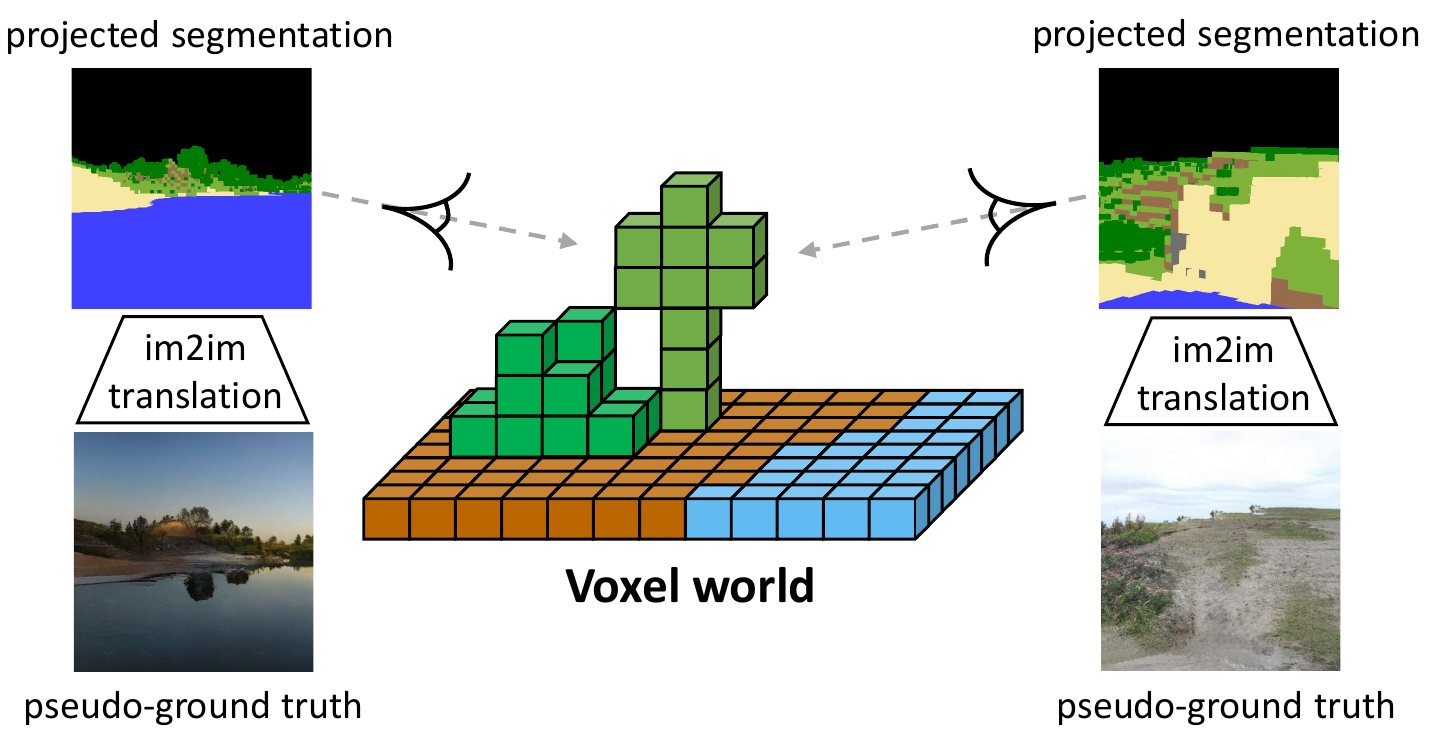}
\end{minipage}%
\begin{minipage}{.54\textwidth}
    \setlength{\tabcolsep}{0.5pt}
    \renewcommand{\arraystretch}{0.2}
    \begin{adjustbox}{max width=\textwidth}
    \begin{tabular}{ccccc}
        \includegraphics[width=0.2\textwidth]{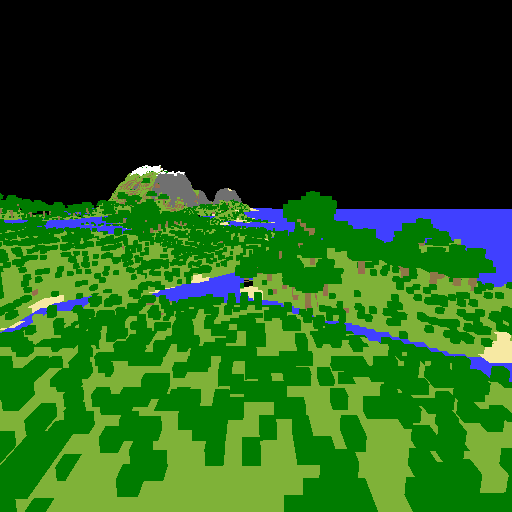} &
        \includegraphics[width=0.2\textwidth]{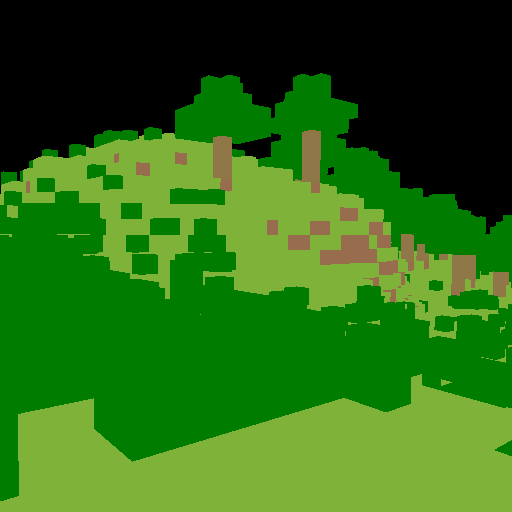} &
        \includegraphics[width=0.2\textwidth]{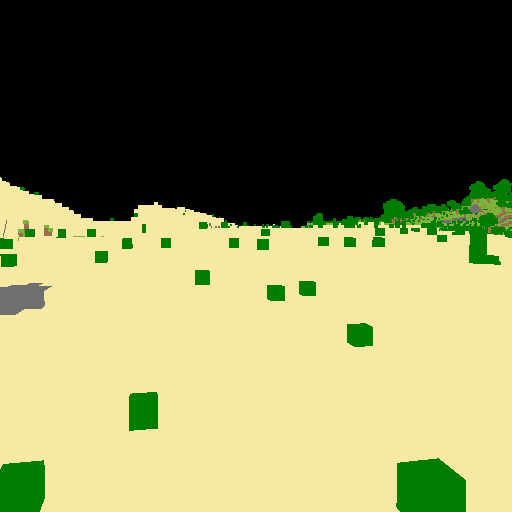} &
        \includegraphics[width=0.2\textwidth]{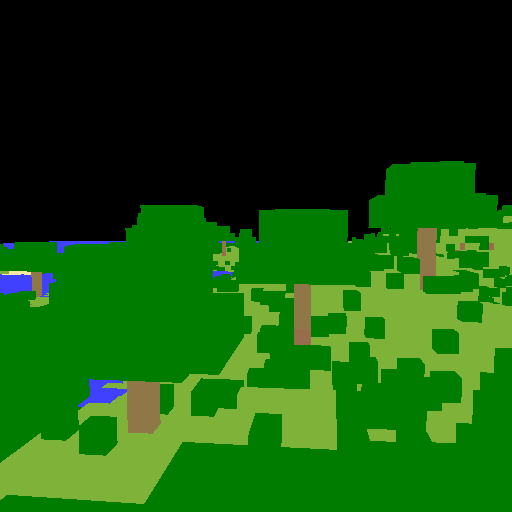} &
        \includegraphics[width=0.2\textwidth]{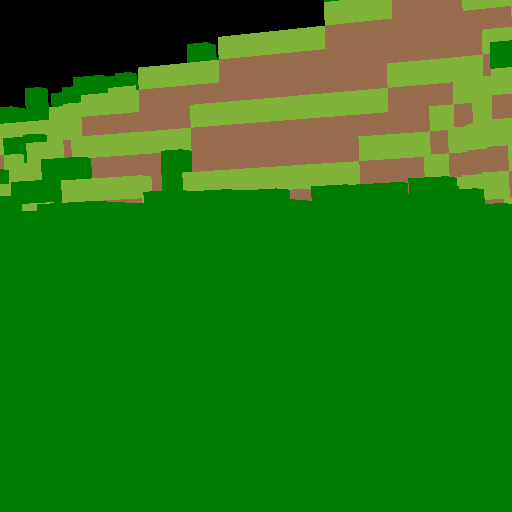} \\
        \includegraphics[width=0.2\textwidth]{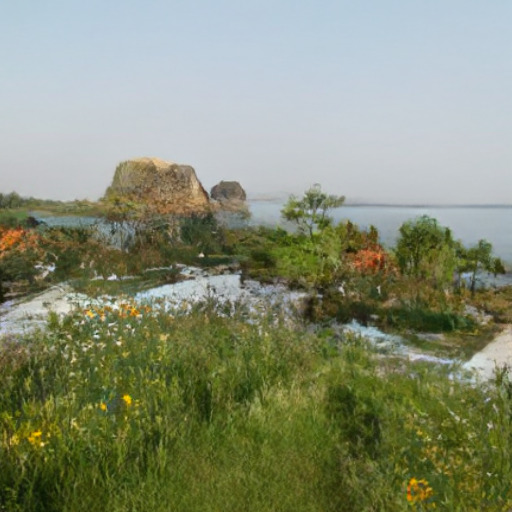} &
        \includegraphics[width=0.2\textwidth]{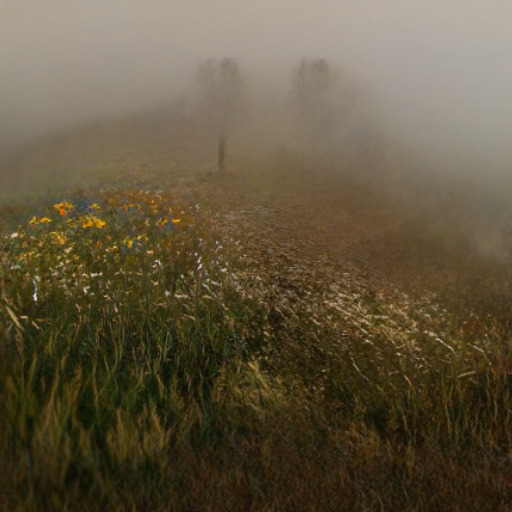} &
        \includegraphics[width=0.2\textwidth]{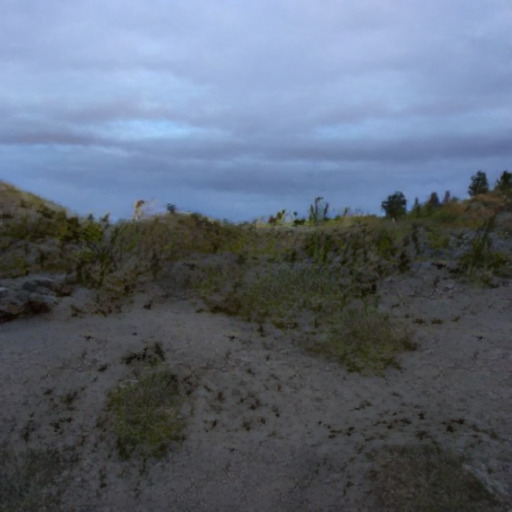} &
        \includegraphics[width=0.2\textwidth]{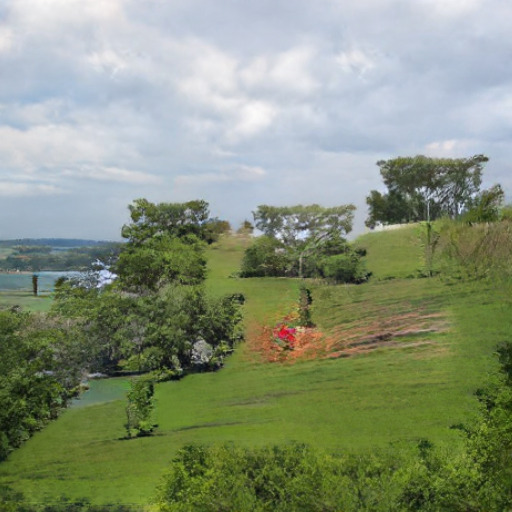} &
        \includegraphics[width=0.2\textwidth]{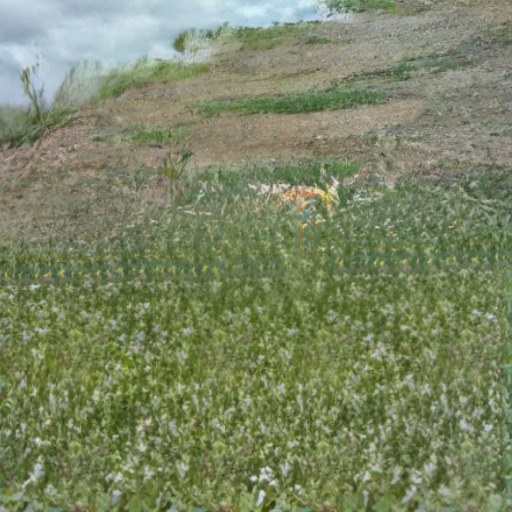}
    \end{tabular}
    \end{adjustbox}
\end{minipage}
\caption{{\bf Pseudo-ground truth generation.} Left: We use a pretrained image-to-image translation model (SPADE~\cite{park2019semantic}) to convert projected segmentation maps to images. Right: Sample input segmentation maps showing different labels (grass, trees, water, sand, sky) and SPADE outputs for different style codes.
Note that some generated outputs can look unrealistic due to domain gap of the blocky segmentations and sampled camera poses, with the real image data used to train SPADE.
Our method is designed to be robust to noise, varying styles, and inconsistencies present in these generated pseudo-ground truth images.}
\label{fig:sample_pseudo_labels}
\end{figure*}

Our goal is to convert a scene represented by semantically-labeled blocks (or voxels), such as the maps from Minecraft, to a photorealistic 3D scene that can be consistently rendered from arbitrary viewpoints (as shown in Fig.~\ref{fig:teaser}). 
In this paper, we focus on landscape scenes that are orders of magnitude larger than single objects or scenes
typically used in the training and evaluation of previous neural rendering works. In all of our experiments, we use voxel grids of 512$\times$512$\times$256 blocks (512$\times$512 blocks horizontally, 256 blocks tall vertically). Given that each Minecraft block is considered to have a size of 1 cubic meter~\cite{minecraft_distance}, each scene covers an area equivalent to 262,144 $m^2$ (65 acres, or the size of 32 soccer fields) in real life. At the same time, our model needs to learn details that are much finer than a single block, such as tree leaves, flowers, and grass, that too without supervision.
As the input voxels and their labels already define the coarse geometry and semantic arrangement of the scene, it is necessary to respect and incorporate this prior information into the model. 
We first describe how we overcome the lack of paired training data by using pseudo-ground truths.
Then, we present our novel sparse voxel-based neural renderer.

\subsection{Generating pseudo-ground truth training data}
\label{subsec:pseudo_gt}
The most straightforward way of training a neural rendering model is to utilize ground truth images with known camera poses. A simple L$_2$ reconstruction loss is sufficient to produce good results in this case~\cite{liu2020neuralsparse,martin2020nerf,mildenhall2020nerf,niemeyer2020differentiable,yariv2020multiview}.
However, in our setting, ground truth real images are simply unavailable for user-generated block worlds from Minecraft. 

An alternative route is to train our model in an unpaired, unsupervised fashion like CycleGAN~\cite{zhu2017unpaired}, or MUNIT~\cite{huang2018multimodal}. This would use an adversarial loss and regularization terms to translate Minecraft segmentations to real images.
However, as shown in the ablation studies in Section~\ref{sec:experiments}, this setting does not produce good results, for both prior methods, and neural renderers. This can be attributed to the large domain gap between blocky Minecraft and the real world, as well as the label distribution shift between worlds.

To bridge the domain gap between the voxel world and our world, we supplement the training data with \textit{pseudo}-ground truth that is generated on-the-fly. 
For each training iteration, we randomly sample camera poses from the upper hemisphere and randomly choose a focal length. We then project the semantic labels of the voxels to the camera view to obtain a 2D semantic segmentation mask. The segmentation mask, as well as a randomly sampled style code, is fed to a pretrained image-to-image translation network, SPADE~\cite{park2019semantic} in our case, to obtain a photorealistic pseudo-ground truth image that has the same semantic layout as the camera view, as shown in the left part of Fig.~\ref{fig:sample_pseudo_labels}. 
This enables us to apply reconstruction losses, such as L$_2$, and the perceptual loss~\cite{johnson2016perceptual}, between the pseudo-ground truth and the rendered output from the same camera view, in additional to the adversarial loss. This significantly improves the result.

The generalizability of the SPADE model trained on large-scale datasets, combined with its photorealistic generation capability helps reduce both, the domain gap and the label distribution mismatch. Sample pseudo-pairs are shown in the right part of Fig.~\ref{fig:sample_pseudo_labels}.
While this provides effective supervision, it is not perfect. This can be seen especially in the last two columns in the right part of Fig.~\ref{fig:sample_pseudo_labels}. The blockiness of Minecraft can produce unrealistic images with sharp geometry. Certain camera poses and style code combinations can also produce images with artifacts. We thus have to be careful to balance reconstruction and adversarial losses to ensure successful training of the neural renderer.

\begin{figure*}[t]
\centering
    \includegraphics[width=\textwidth, trim={0.05in, 4.84in, 0.60in, 1.15in}, clip]{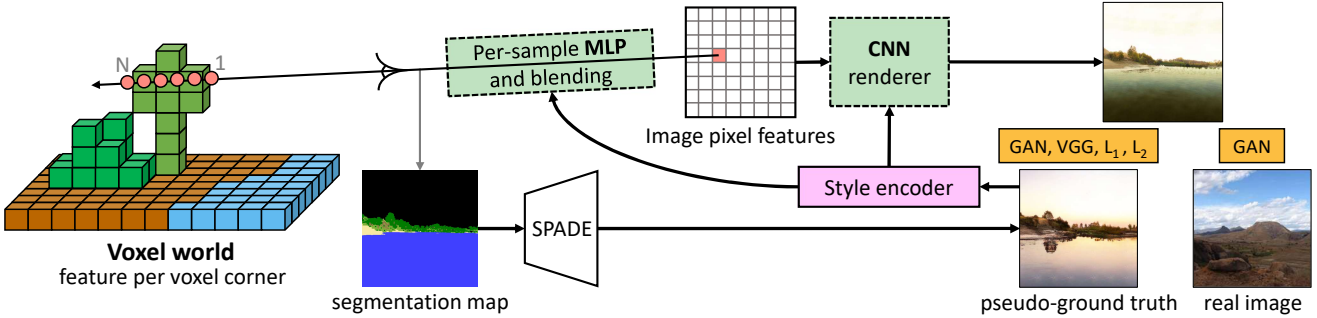}
    \caption{{\bf Overview of GANcraft.} Given an input voxel world with segmentation labels, we first assign features to every voxel corner. For arbitrarily sampled camera viewpoints, we obtain the trilinearly interpolated voxel features at the point of ray-voxel intersections, process with an MLP, and blend the output features to obtain the image pixel features.
    These features are fed to an image-space CNN renderer. Both the MLP and the CNN are conditioned on the style code of the pseudo-ground truth for the chosen camera view.
    Our method is trained with an adversarial loss with real images, and a combination of adversarial, pixel-wise, and VGG perceptual losses on the pseudo-ground truths. After training, we can render the world in a photorealistic manner, controlling the style of the output images by conditioning on an input style code or image.}
    \label{fig:overview}
\end{figure*}
\subsection{Sparse voxel-based volumetric neural renderer}
\label{subsec:network_arch}
\noindent{\bf Voxel-bounded neural radiance fields.} 
Let $K$ be the number of occupied blocks in a Minecraft world, which can also be represented by a sparse voxel grid with $K$ non-empty voxels given by $\mathcal{V} = \{V_1, ..., V_K\}$. Each voxel is assigned a semantic label $\{l_1,...,l_K\}$. We learn a neural radiance field per voxel. The Minecraft world is then represented by the union of voxel-bounded neural radiance fields given by
\begin{gather}
    F(\bp, \bz) = \begin{cases}
        F_i(\bp, \bz), &~\text{if } \bp \in V_i, ~\ i\in\{1, \cdots, K\}\\
        (\mathbf{0}, 0), &~\text{otherwise}
    \end{cases}, %
\end{gather}
where $F$ is the radiance field of the whole scene and $F_i$ is the radiance field bounded by $V_i$. Querying a location in the neural radiance field returns a feature vector (or color in prior work~\cite{liu2020neuralsparse,martin2020nerf,mildenhall2020nerf}) and a density value. At the location where a block does not exist, we have the null feature vector $\mathbf{0}$ and zero density $0$. To model diversified appearance of the same scene, \eg day and night, the radiance fields are conditioned on style code $\bz$. 

The voxel-bounded neural radiance field $F_i$ is given by 
\begin{equation}
F_i(\bp, \bz)= G_\theta(g_i(\bp), l_i, \bz) = \left(\bc(\bp, l(\bp), \bz), \sigma(\bp, l(\bp))\right),\nonumber
\end{equation}
where $g_i(\bp)$ is the location code at $\bp$ and $l_i\equiv l(\bp)$ is a short-hand for the label of the voxel that $\bp$ belongs to. The multi-layer perceptron (MLP) $G_\theta$ is used to predict the feature $\bc$, and volume density $\sigma$ at the location~$\bp$. We note that $G_\theta$ is shared amongst all voxels. Inspired by NeRF-W~\cite{martin2020nerf}, $\bc$ additionally depends on the style code, while the density $\sigma$ does not. To obtain the location code, we first assign a learnable feature vector to each of the eight vertices of a voxel $V_i$. The location code at $\bp$, $g_i(\bp)$, is then derived through trilinear interpolation. 
Here, we assume that each voxel has a shape of 1$\times$1$\times$1, and the coordinate axes are aligned to the voxel grid axes. Vertices and their feature vectors are shared for adjacent voxels. This allows for a smooth transition of features when crossing the voxel boundaries, preventing discontinuities in the output. 
We compute Fourier features from $g_i(\bp)$, similar to NSVF~\cite{liu2020neuralsparse}, and also append the voxel class label. Our method can be interpreted as a generalization of NSVF~\cite{liu2020neuralsparse} to use a style and semantic label conditioning. 

\medskip
\noindent{\bf Neural sky dome.} The sky is an indispensable part of photorealistic landscape scenes. However, as it is physically located much farther away from the other objects, it is inefficient to represent it with a layer of voxels. In GANcraft, we assume that the sky is located infinitely far away (no parallax). Thus, its appearance is only dependent on the viewing direction. The same assumption is commonly used in computer graphics techniques such as environment mapping~\cite{blinn1976texture}. We use an MLP $H_\phi$, to map ray direction $\bv$ to sky color, or feature, 
$\bc_\mathrm{sky} \equiv H_\phi(\bv, \bz)$,
conditioned on style code $\bz$,
This representation can be viewed as covering the whole scene with an infinitely large sky dome.

\medskip
\noindent{\bf Volumetric rendering.}
Here, we describe how a scene represented by the above-mentioned neural radiance fields and sky dome can be converted to 2D feature maps via volumetric rendering. Under a perspective camera model, each pixel in the image corresponds to a camera ray $\br(t) = \bo + t \bv$, originating from the center of projection $\bo$ and advances in direction $\bv$. The ray travels through the radiance field while accumulating features and transmittance,
\begin{align}
    &C(\br, \bz) = \int_0^{+\infty} T(t) \sigma\big{(}\br(t), l(\br(t))\big{)} \bc\big{(}\br(t), l(\br(t)), \bz\big{)} dt \nonumber \label{eq:vol_rend_color} \\
    & \quad\quad\quad\quad\ + T(+\infty) \bc_\mathrm{sky}(\bv, \bz),\\
    &\mathrm{where}~ T(t) = \exp{\left( - \int_0^t \sigma(\br(s))ds \right)}.
\end{align}
$C(\br, \bz)$ denotes the accumulated feature of ray $\br$, and $T(t)$ denotes the accumulated transmittance when the ray travels a distance of $t$. 
As the radiance field is bounded by a finite number of voxels, the ray will eventually exit the voxels and hit the sky dome. We thus consider the sky dome as the last data point on the ray, which is totally opaque. This is realized by the last term in Eq.~\ref{eq:vol_rend_color}.
The above integral can be approximated using discrete samples and the quadrature rule, a technique popularized by NeRF~\cite{mildenhall2020nerf}. Please refer to NeRF~\cite{mildenhall2020nerf} or our supplementary for the full equations.

We use the stratified sampling technique from NSVF~\cite{liu2020neuralsparse} to randomly sample valid (voxel bounded) points along the ray. To improve efficiency, we truncate the ray so that it will stop after a certain accumulated distance through the valid region is reached. We regularize the truncated rays to encourage their accumulated opacities to saturate before reaching the maximum distance. 
We adopt a modified Bresenham method \cite{amanatides1987fast} for sampling valid points, which has a very low complexity of $O(N)$, where $N$ is the longest dimension of the voxel grid. Details are in the supplementary.

\medskip
\noindent{\bf Hybrid neural rendering architecture.} Prior works~\cite{liu2020neuralsparse,martin2020nerf,mildenhall2020nerf} directly produce images by accumulating colors using the volumetric rendering scheme described above instead of accumulating features. Unlike them, we divide rendering into two parts: 1) We perform volumetric rendering with an MLP to produce a feature vector per pixel instead of an RGB image, and 2) We employ a CNN to convert the per-pixel feature map to a final RGB image of the same size. The overall framework is shown in Fig.~\ref{fig:overview}. We perform activation modulation~\cite{huang2017adain,park2019semantic} conditioned on the input style code for both the MLP and CNN. The individual networks are described in detail in the supplementary.

Apart from improving the output image quality as shown in Section~\ref{sec:experiments}, this two-stage design also helps reduce the computational and memory footprint of rendering. The MLP modeling the 3D radiance field is evaluated on a per-sample basis, while the image-space CNN is only evaluated after multiple samples along a ray are merged into a single pixel. The number of samples to the MLP scales linearly with the output height, width, and number of points sampled per ray (24 in our case), while the size of the feature map only depends on output height and width. However, unlike MLPs that operate pre-blending, the image-space CNN 
is not intrinsically view consistent. We thus use a shallow CNN with a receptive field of only 9$\times$9 pixels to constrain its scope to local manipulations.
A similar idea of combining volumetric rendering and image-space rendering has been used in GIRAFFE~\cite{niemeyer2020giraffe}. Unlike us, they also rely on the CNN to upsample a low-resolution 16$\times$16 feature map.

\medskip
\noindent{\bf Losses and regularizers.}
We train our model with both reconstruction and adversarial losses.
The reconstruction loss is applied between the predicted images and the corresponding pseudo-ground truth images. We use a combination of the perceptual~\cite{johnson2016perceptual}, L$_1$, and L$_2$ losses.
For the GAN loss, we treat the predicted images as `fake', and both the real images, and the pseudo-ground truth images as `real'.  We use a discriminator conditioned on the semantic segmentation maps, based on Liu~\etal~\cite{liu2019learning} and Sch{\"o}nfeld~\etal~\cite{schonfeld2021you}.  We use the hinge loss~\cite{lim2017geometric} as the GAN training objective. Following previous works on multimodal image synthesis~\cite{gaugan,huang2018multimodal,zhu2017toward}, we also include a style encoder which predicts the posterior distribution of the style code given a pseudo-ground truth image. The reconstruction loss, in conjunction with the style encoder, makes it possible to control the appearance of the output image with a style image. 

As mentioned earlier, we truncate the ray during volumetric rendering. To avoid artifacts due to the truncation, we apply an opacity regularization term on the truncated ray, $\mathcal{L}_\mathrm{opacity} = \sum_{\br \in \mathbf{R_\mathrm{trunc}}} T_\textrm{out}(\br)$.
This discourages leftover transmittance after a ray reaches the truncation distance.

\section{Experiments}
\label{sec:experiments}
The previous section described how we obtain pseudo-ground truths in the absence of paired Minecraft--real training data, and the architecture of our neural renderer. Here, we validate our framework by comparing with prior work on multiple diverse large Minecraft worlds.

\noindent{\bf Datasets.} We collected a dataset of $\sim$1M landscape images from the internet with a minimum side of at least 512 pixels. For each image, we obtained 182-class COCO-Stuff~\cite{caesar2018coco} segmentation labels by using DeepLabV2~\cite{chen2017deeplab,nakashima_deeplab}. This formed our training set of paired real segmentation maps and images. We set aside 5000 images as a test set.
We generated 5 different Minecraft worlds of 512$\times$512$\times$256 blocks each. We sampled worlds with various compositions of water, sand, forests, and snow, to show that our method works correctly under significant label distribution shifts.

\smallskip
\noindent{\bf Baselines.} We compare against the following, which are representative methods under different data availability regimes.
\begin{itemize}[leftmargin=11pt, label=\textbullet, topsep=2pt, itemsep=-3pt]
    \item MUNIT~\cite{huang2018multimodal}. This is an image-to-image translation method trainable in the unpaired, or unsupervised setting. Unlike CycleGAN~\cite{zhu2017unpaired} and UNIT~\cite{liu2016unsupervised}, MUNIT can learn multimodal translations. We learn to translate Minecraft segmentation maps to real images.
    \item SPADE~\cite{park2019semantic}. This is an image-to-image translation method that is trained in the paired ground truth, or supervised setting.
    We train this by translating real segmentation maps to corresponding images and test it on Minecraft segmentations.
    \item wc-vid2vid~\cite{mallya2020world}. Unlike the above two methods, this can generate a sequence of images that are view-consistent. wc-vid2vid projects the pixels from previous frames to the next frame 
    to generate a guidance map. This serves as a form of memory of the previously generated frames. This method also requires paired ground truth data, as well as the 3D point clouds for each output frame. We train this to translate real segmentation maps to real images, while using the block world voxel surfaces as the 3D geometry.
    \item NSVF-W~\cite{liu2020neuralsparse,martin2020nerf}. We combine the strengths of two recent works on neural rendering, NSVF~\cite{liu2020neuralsparse}, and NeRF-W~\cite{martin2020nerf}, to create a strong baseline. NSVF represents the world as voxel-bounded radiance fields, and can be modified to accept an input voxel world, just like our method. NeRF-W is able to learn from unstructured image collections with variations in color, lighting, and occlusions, making it well-suited for learning from our pseudo-ground truths. Combining the style-conditioned MLP generator from NeRF-W with the voxel-based input representation of NSVF, we obtain NSVF-W. This resembles the neural renderer used by us, with the omission of the image-space CNN. As these methods also require paired ground truth, we train NSVF-W using pseudo-ground truths generated by the pretrained SPADE model.
\end{itemize}
MUNIT, SPADE, and wc-vid2vid use perceptual and adversarial losses during training, while NSVF, NeRF-W, and thus NSVF-W use the L$_2$ loss. Details are in the supplementary.

\smallskip
\noindent{\bf Implementation details.}
We train our model at an output resolution of 256$\times$256 pixels.
Each model is trained on 8 NVIDIA V100 GPUs with 32GB of memory each. This enables us to use a batch size of 8 with 24 points sampled per camera ray. Each model is trained for 250k iterations, which takes approximately 4 days. All baselines are also trained for an equivalent amount of time. Additional details are available in the supplementary.

\smallskip
\noindent{\bf Evaluation metrics.} We use both quantitative and qualitative metrics to measure the quality of our outputs.
\begin{itemize}[leftmargin=11pt, label=\textbullet, topsep=2pt, itemsep=-3pt]
    \item Fr\'echet Inception Distance~\cite{heusel2017gans} (FID) and Kernel Inception Distance~\cite{binkowski2018demystifying} (KID).
    We use FID and KID to measure the distance between the distributions of the generated and real images, using Inception-v3~\cite{szegedy2016rethinking}. We generate 1000 images for each of the 5 worlds from arbitrarily sampled camera view points using different style codes, for a total of 5000 images. We then generate outputs from each method for the same pair of view points and style code for a fair comparison. We use a held-out set of 5000 real landscape images to compute the metrics. For both metrics, a lower value indicates better image quality.
    \item Human preference score. Using Amazon Mechanical Turk (AMT), we perform a subjective visual test to gauge the relative quality of generated videos with top-qualified turkers. We ask turkers to choose 1) the more temporally consistent video, and 2) the video with overall better realism. For each of the two questions, a turker is shown two videos synthesized by two different methods and asked to choose the superior one according to the criteria. We generate 64 videos per world, total of 320 per method, and each comparison is evaluated by 3 workers.
\end{itemize}

\begin{table}[t]
    \centering
    \begin{tabular}{lcc}
        \toprule
        Method & FID $\downarrow$ & KID $\downarrow$ \\
        \midrule
        SPADE~\cite{park2019semantic} & 58.90 & 0.027 \\
        \midrule
        MUNIT~\cite{huang2018multimodal} & 78.42 & 0.047 \\
        NSVF-W~\cite{liu2020neuralsparse,martin2020nerf} & 84.53 & 0.052 \\
        GANcraft (ours) & {\bf 61.33} & {\bf 0.033} \\
        \bottomrule
    \end{tabular}
    \caption{{\bf Automated image quality metrics.} We compare baselines against GANcraft on all 5 block worlds.
    SPADE sets the lower bound on FID and KID as it is a strong photorealistic image generator, although it is not view-consistent. Our view-consistent method achieves values close to SPADE, beating MUNIT and NSVF-W.}
    \label{table:metrics}
\end{table}

\smallskip
\noindent{\bf Main results.}
Fig.~\ref{fig:view_consistent outputs} shows output videos generated by different methods. Each row is a unique world, generated using the same style-conditioning image for all methods. We can observe that our outputs are more realistic and view-consistent when compared to baselines. MUNIT~\cite{huang2018multimodal} and SPADE~\cite{park2019semantic} demonstrate a lot of flickering as they generate one image at a time, without any memory of past outputs. Further, MUNIT also fails to learn the correct mapping of segmentation labels to textures as it does not use paired supervision.
While wc-vid2vid~\cite{mallya2020world} is more view-consistent, it fails for large motions as it incrementally inpaints newly explored parts of the world. NSVF-W~\cite{liu2020neuralsparse,martin2020nerf} and GANcraft are both inherently view-consistent due to their use of volumetric rendering. However, due to the lack of a CNN renderer and the use of the L$_2$ loss, NSVF-W produces dull and unrealistic outputs with artifacts. The use of an adversarial loss is key to ensuring vivid and realistic results, and this is further reinforced in the ablations presented below. Our method is also capable of generating higher resolution outputs as shown in Fig.~\ref{fig:teaser}, by sampling more rays.

We sample novel camera views from each world and compute the FID and KID against a set of held-out real images. As seen in Table~\ref{table:metrics}, our method achieves FID and KID close to that of SPADE, which is a very strong image-to-image translation method, while beating other baselines. Note that wc-vid2vid uses SPADE to generate the output for first camera view in a sequence and is thus ignored in this comparison. Further, as summarized in Table~\ref{table:user_study}, users consistently preferred our method and chose its predictions as the more view-consistent and realistic videos. More high-resolution results and comparisons as well as some failure cases are available in the supplementary.

\begin{table}[t]
    \centering
    \begin{adjustbox}{max width=\columnwidth}
    \begin{tabular}{r@{\hskip2pt}c@{\hskip2pt}lcc}
        \toprule
        \multicolumn{3}{c}{\multirow{2}{*}{Comparison}} & \multicolumn{2}{c}{Human preference} \\
        & & & Consistency $\uparrow$ & Realism $\uparrow$ \\
        \midrule
        MUNIT~\cite{huang2018multimodal} & /& GANcraft & 30.1/{\bf 69.9} & 37.5/{\bf 62.5} \\
        SPADE~\cite{park2019semantic} & /& GANcraft & 29.7/{\bf 70.3} & 37.2/{\bf 62.8}  \\
        wc-vid2vid~\cite{mallya2020world} & /& GANcraft & 47.0/{\bf 53.0} & 16.2/{\bf 83.8} \\
        NSVF-W~\cite{liu2020neuralsparse,martin2020nerf} & /& GANcraft & 46.6/{\bf 53.4} & 31.4/{\bf 68.6} \\
        \bottomrule
    \end{tabular}
    \end{adjustbox}
    \caption{{\bf Human preference scores.} We compare videos generated by different methods on all 5 block worlds. Users chose GANcraft as more temporally consistent and realistic. }
    \label{table:user_study}
\end{table}

\begin{table*}[t]
    \centering
    \setlength{\tabcolsep}{0pt}
    \begin{adjustbox}{max width=\textwidth}
        \begin{tabular}{C{0.2\textwidth}C{0.2\textwidth}C{0.2\textwidth}C{0.2\textwidth}C{0.2\textwidth}}
            MUNIT~\cite{huang2018multimodal} &
            SPADE~\cite{park2019semantic} &
            wc-vid2vid~\cite{mallya2020world} &
            NSVF-W~\cite{liu2020neuralsparse,martin2020nerf} &
            {\bf GANcraft (ours)} \\
            \multicolumn{5}{c}{\href{https://nvlabs.github.io/GANcraft/videos/desert_2527_2.mp4}{\includegraphics[width=\textwidth]{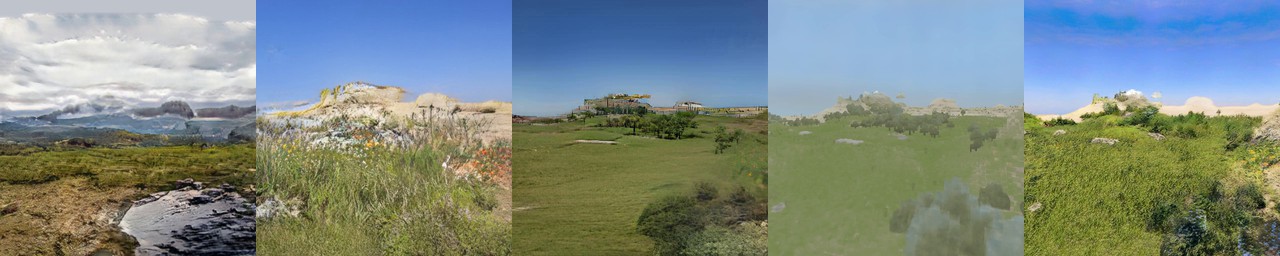}}} \\[-4pt]
            \multicolumn{5}{c}{\href{https://nvlabs.github.io/GANcraft/videos/survivalisland_2467_1.mp4}{\includegraphics[width=\textwidth]{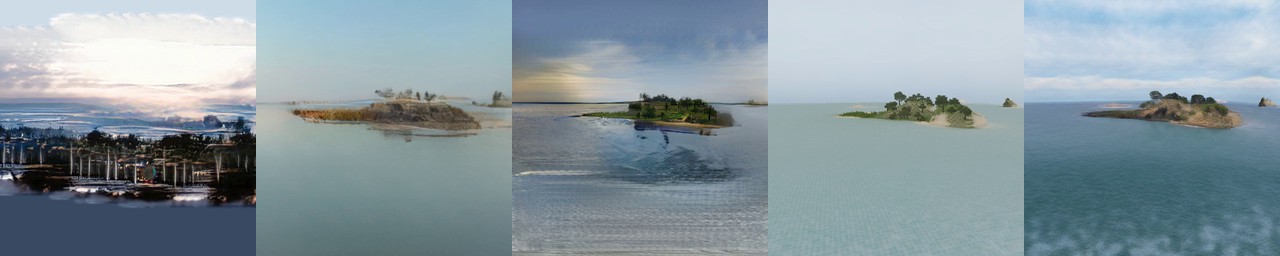}}} \\[-4pt]
            \multicolumn{5}{c}{\href{https://nvlabs.github.io/GANcraft/videos/landscapep2_0685_3.mp4}{\includegraphics[width=\textwidth]{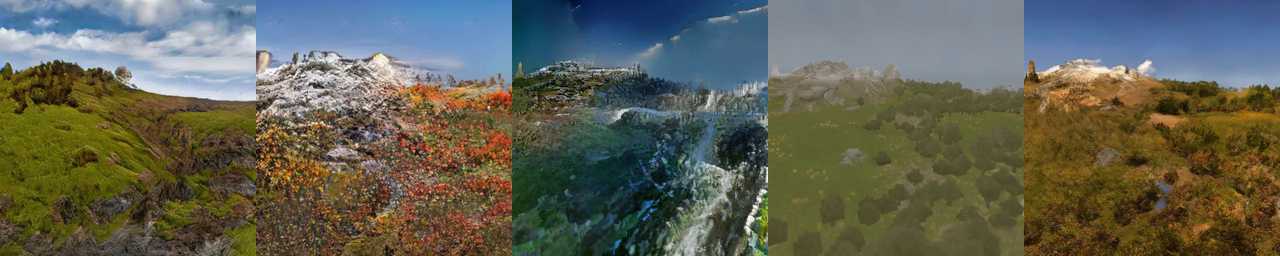}}} \\[-4pt]
            \multicolumn{5}{c}{\href{https://nvlabs.github.io/GANcraft/videos/s123456_0298_1.mp4}{\includegraphics[width=\textwidth]{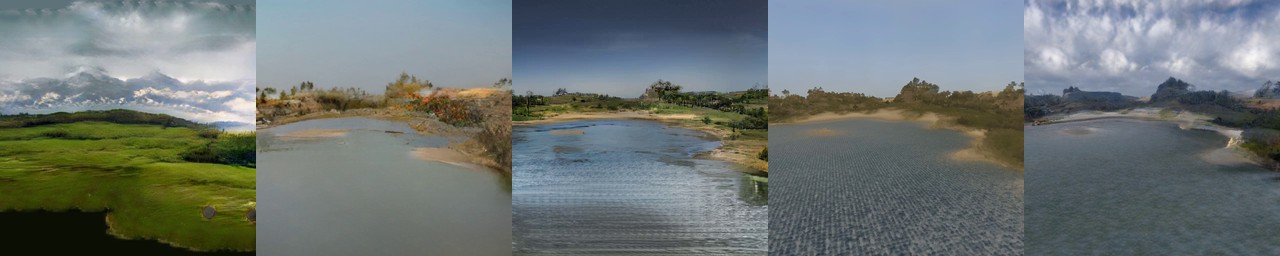}}}
        \end{tabular}
    \end{adjustbox}
    \captionof{figure}{{\bf Output video comparison.} Each row is a unique world, and each column is a different method. For a given world, all methods use the same style-conditioning image. GANcraft produces more view-consistent and more realistic outputs compared to all baselines.
    \emph{Click any row to play video in web browser.} \acrobat}
    \label{fig:view_consistent outputs}
\end{table*}

\begin{figure*}[t]
\vspace{4pt}
\centering
\setlength{\tabcolsep}{0pt}
    \begin{adjustbox}{max width=\textwidth}
        \begin{tabular}{C{0.2\textwidth}C{0.2\textwidth}C{0.2\textwidth}C{0.2\textwidth}C{0.2\textwidth}}
            Full model & No pseudo-ground truth & No CNN & No real images & No GAN loss \\
            \multicolumn{5}{c}{\includegraphics[width=\textwidth]{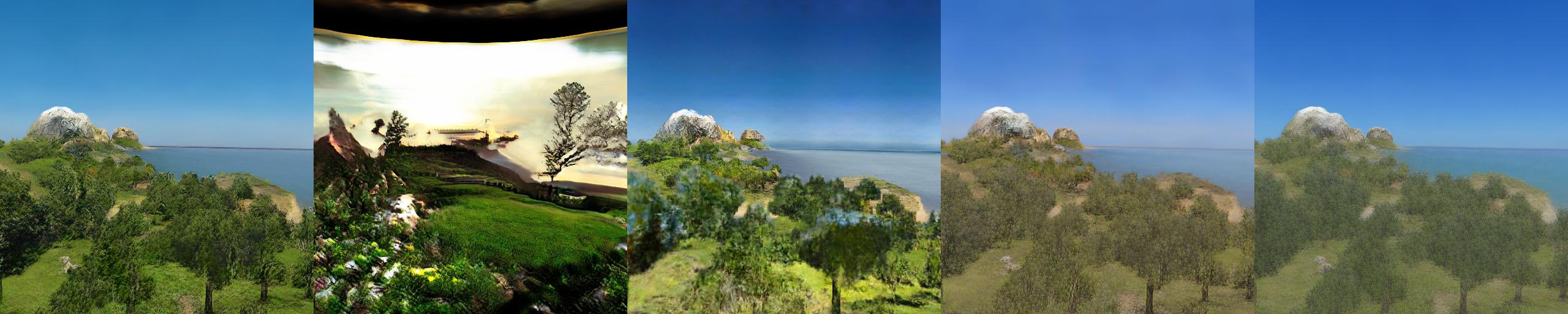}}
        \end{tabular}
    \end{adjustbox}
    \caption{{\bf Ablated model outputs.} Using only the GAN loss with no pseudo-ground truths produces unrealistic images. Not using a CNN produces outputs that lack detail and contain artifacts. Excluding the GAN loss on real images results in dull colors, and no GAN loss at all produces dull and blurry outputs, when compared to the full model.}
    \label{fig:ablation}
\end{figure*}
\smallskip
\noindent{\bf Ablations.}
We train ablated versions of our full model on one Minecraft world due to computational constraints. 
We show example outputs from them in Fig.~\ref{fig:ablation}. Using no pseudo-ground truth at all and training with just the GAN loss produces unrealistic outputs, similar to MUNIT~\cite{huang2018multimodal}. Directly producing images from volumetric rendering, 
without using a CNN, results in a lack of fine detail. Compared to the full model, skipping the GAN loss on real images produces duller images, and skipping the GAN loss altogether produces duller, blurrier images resembling NSVF-W outputs. Qualitative analysis is available in the supplementary.

\section{Discussion}
\label{sec:discussion}
We introduced the novel task of \emph{world-to-world} translation and proposed GANcraft, a method to convert block worlds to realistic-looking worlds. We showed that \emph{pseudo}-ground truths generated by a 2D image-to-image translation network provide effective means of supervision in the absence of real paired data. Our hybrid neural renderer trained with both real landscape images and pseudo-ground truths, and adversarial losses, outperformed strong baselines.

There still remain a few exciting avenues for improvements, including learning a smoother geometry in spite of coarse input geometry, 
and using non-voxel inputs such as meshes and point clouds. 
While our method is currently trained on a per-world basis, we hope future work can enable feed-forward generation on novel worlds.

\noindent{\bf Acknowledgements.} We thank Rev Lebaredian for challenging us to work on this interesting problem. We thank Jan Kautz, Sanja Fidler, Ting-Chun Wang, Xun Huang, Xihui Liu, Guandao Yang, and Eric Haines for their feedback during the course of developing the method.

{\small
\bibliographystyle{ieee_fullname}
\bibliography{egbib}

\begin{thebibliography}{10}\itemsep=-1pt

\bibitem{minecraft_distance}
{Minecraft units of measure}.
\newblock
  \url{https://minecraft.gamepedia.com/Tutorials/Units_of_measure\#Distance}.

\bibitem{minecraft_user_stats}
{Minecraft user statistics}.
\newblock \url{https://bit.ly/3sZkaMF}.

\bibitem{gaugan}
{NVIDIA GauGAN}.
\newblock \url{https://blogs.nvidia.com/blog/2019/07/30/gaugan-ai-painting/}.

\bibitem{aliev2019neural}
Kara-Ali Aliev, Artem Sevastopolsky, Maria Kolos, Dmitry Ulyanov, and Victor
  Lempitsky.
\newblock Neural point-based graphics.
\newblock In {\em ECCV}, 2020.

\bibitem{amanatides1987fast}
John Amanatides and Andrew Woo.
\newblock A fast voxel traversal algorithm for ray tracing.
\newblock In {\em Eurographics}, 1987.

\bibitem{bi2020neural}
Sai Bi, Zexiang Xu, Pratul Srinivasan, Ben Mildenhall, Kalyan Sunkavalli,
  Milo{\v{s}} Ha{\v{s}}an, Yannick Hold-Geoffroy, David Kriegman, and Ravi
  Ramamoorthi.
\newblock Neural reflectance fields for appearance acquisition.
\newblock {\em arXiv preprint arXiv:2008.03824}, 2020.

\bibitem{binkowski2018demystifying}
Mikołaj Bińkowski, Dougal~J. Sutherland, Michael Arbel, and Arthur Gretton.
\newblock Demystifying {MMD} {GAN}s.
\newblock In {\em ICLR}, 2018.

\bibitem{blinn1976texture}
James~F Blinn and Martin~E Newell.
\newblock Texture and reflection in computer generated images.
\newblock {\em Comm. of the ACM}, 1976.

\bibitem{boss2020nerd}
Mark Boss, Raphael Braun, Varun Jampani, Jonathan~T Barron, Ce Liu, and Hendrik
  Lensch.
\newblock {NeRD}: Neural reflectance decomposition from image collections.
\newblock {\em arXiv preprint arXiv:2012.03918}, 2020.

\bibitem{caesar2018coco}
Holger Caesar, Jasper Uijlings, and Vittorio Ferrari.
\newblock {COCO-Stuff}: Thing and stuff classes in context.
\newblock In {\em CVPR}, 2018.

\bibitem{chan2020pi}
Eric~R Chan, Marco Monteiro, Petr Kellnhofer, Jiajun Wu, and Gordon Wetzstein.
\newblock {pi-GAN}: Periodic implicit generative adversarial networks for
  3d-aware image synthesis.
\newblock In {\em CVPR}, 2021.

\bibitem{chen2017deeplab}
Liang-Chieh Chen, George Papandreou, Iasonas Kokkinos, Kevin Murphy, and Alan~L
  Yuille.
\newblock {DeepLab}: Semantic image segmentation with deep convolutional nets,
  atrous convolution, and fully connected crfs.
\newblock {\em IEEE TPAMI}, 2017.

\bibitem{choi2019pseudo}
Jaehoon Choi, Minki Jeong, Taekyung Kim, and Changick Kim.
\newblock Pseudo-labeling curriculum for unsupervised domain adaptation.
\newblock In {\em BMVC}, 2019.

\bibitem{choi2017stargan}
Yunjey Choi, Minje Choi, Munyoung Kim, Jung-Woo Ha, Sunghun Kim, and Jaegul
  Choo.
\newblock {StarGAN}: Unified generative adversarial networks for multi-domain
  image-to-image translation.
\newblock In {\em CVPR}, 2018.

\bibitem{du2020neural}
Yilun Du, Yinan Zhang, Hong-Xing Yu, Joshua~B. Tenenbaum, and Jiajun Wu.
\newblock Neural radiance flow for 4d view synthesis and video processing.
\newblock {\em arXiv preprint arXiv:2012.09790}, 2020.

\bibitem{goodfellow2014generative}
Ian Goodfellow, Jean Pouget-Abadie, Mehdi Mirza, Bing Xu, David Warde-Farley,
  Sherjil Ozair, Aaron Courville, and Yoshua Bengio.
\newblock Generative adversarial networks.
\newblock In {\em NeurIPS}, 2014.

\bibitem{guo2020object}
Michelle Guo, Alireza Fathi, Jiajun Wu, and Thomas Funkhouser.
\newblock Object-centric neural scene rendering.
\newblock {\em arXiv preprint arXiv:2012.08503}, 2020.

\bibitem{henzler2019escaping}
Philipp Henzler, Niloy~J Mitra, and Tobias Ritschel.
\newblock Escaping {P}lato's cave: 3d shape from adversarial rendering.
\newblock In {\em CVPR}, 2019.

\bibitem{heusel2017gans}
Martin Heusel, Hubert Ramsauer, Thomas Unterthiner, Bernhard Nessler, and Sepp
  Hochreiter.
\newblock {GANs} trained by a two time-scale update rule converge to a local
  {Nash} equilibrium.
\newblock In {\em NeurIPS}, 2017.

\bibitem{huang2017adain}
Xun Huang and Serge Belongie.
\newblock Arbitrary style transfer in real-time with adaptive instance
  normalization.
\newblock In {\em ICCV}, 2017.

\bibitem{huang2018multimodal}
Xun Huang, Ming-Yu Liu, Serge Belongie, and Jan Kautz.
\newblock Multimodal unsupervised image-to-image translation.
\newblock {\em ECCV}, 2018.

\bibitem{isola2017image}
Phillip Isola, Jun-Yan Zhu, Tinghui Zhou, and Alexei~A Efros.
\newblock Image-to-image translation with conditional adversarial networks.
\newblock In {\em CVPR}, 2017.

\bibitem{johnson2016perceptual}
Justin Johnson, Alexandre Alahi, and Li Fei-Fei.
\newblock Perceptual losses for real-time style transfer and super-resolution.
\newblock In {\em ECCV}, 2016.

\bibitem{karras2018style}
Tero Karras, Samuli Laine, and Timo Aila.
\newblock A style-based generator architecture for generative adversarial
  networks.
\newblock In {\em CVPR}, 2019.

\bibitem{karras2020analyzing}
Tero Karras, Samuli Laine, Miika Aittala, Janne Hellsten, Jaakko Lehtinen, and
  Timo Aila.
\newblock Analyzing and improving the image quality of {StyleGAN}.
\newblock In {\em CVPR}, 2020.

\bibitem{lee2020maskgan}
Cheng-Han Lee, Ziwei Liu, Lingyun Wu, and Ping Luo.
\newblock {MaskGAN}: Towards diverse and interactive facial image manipulation.
\newblock In {\em CVPR}, 2020.

\bibitem{lee2013pseudo}
Dong-Hyun Lee.
\newblock Pseudo-label: The simple and efficient semi-supervised learning
  method for deep neural networks.
\newblock In {\em Workshop on challenges in representation learning, ICML},
  2013.

\bibitem{li2020neural}
Zhengqi Li, Simon Niklaus, Noah Snavely, and Oliver Wang.
\newblock Neural scene flow fields for space-time view synthesis of dynamic
  scenes.
\newblock In {\em CVPR}, 2021.

\bibitem{lim2017geometric}
Jae~Hyun Lim and Jong~Chul Ye.
\newblock Geometric {GAN}.
\newblock {\em arXiv preprint arXiv:1705.02894}, 2017.

\bibitem{lindell2020autoint}
David~B Lindell, Julien~NP Martel, and Gordon Wetzstein.
\newblock {AutoInt}: Automatic integration for fast neural volume rendering.
\newblock {\em arXiv preprint arXiv:2012.01714}, 2020.

\bibitem{liu2020neuralsparse}
Lingjie Liu, Jiatao Gu, Kyaw~Zaw Lin, Tat-Seng Chua, and Christian Theobalt.
\newblock Neural sparse voxel fields.
\newblock In {\em NeurIPS}, 2020.

\bibitem{liu2016unsupervised}
Ming-Yu Liu, Thomas Breuel, and Jan Kautz.
\newblock Unsupervised image-to-image translation networks.
\newblock In {\em NeurIPS}, 2017.

\bibitem{liu2019few}
Ming-Yu Liu, Xun Huang, Arun Mallya, Tero Karras, Timo Aila, Jaakko Lehtinen,
  and Jan Kautz.
\newblock Few-shot unsupervised image-to-image translation.
\newblock In {\em ICCV}, 2019.

\bibitem{liu2020generative}
Ming-Yu Liu, Xun Huang, Jiahui Yu, Ting-Chun Wang, and Arun Mallya.
\newblock Generative adversarial networks for image and video synthesis:
  Algorithms and applications.
\newblock {\em Proc. of the IEEE}, 2021.

\bibitem{liu2019learning}
Xihui Liu, Guojun Yin, Jing Shao, Xiaogang Wang, et~al.
\newblock Learning to predict layout-to-image conditional convolutions for
  semantic image synthesis.
\newblock In {\em NeurIPS}, 2019.

\bibitem{mallya2020world}
Arun Mallya, Ting-Chun Wang, Karan Sapra, and Ming-Yu Liu.
\newblock World-consistent video-to-video synthesis.
\newblock In {\em ECCV}, 2020.

\bibitem{martin2020nerf}
Ricardo Martin-Brualla, Noha Radwan, Mehdi~SM Sajjadi, Jonathan~T Barron,
  Alexey Dosovitskiy, and Daniel Duckworth.
\newblock {NeRF in the Wild}: Neural radiance fields for unconstrained photo
  collections.
\newblock In {\em CVPR}, 2021.

\bibitem{mcclosky2006effective}
David McClosky, Eugene Charniak, and Mark Johnson.
\newblock Effective self-training for parsing.
\newblock In {\em NAACL}, 2006.

\bibitem{mildenhall2020nerf}
Ben Mildenhall, Pratul~P. Srinivasan, Matthew Tancik, Jonathan~T. Barron, Ravi
  Ramamoorthi, and Ren Ng.
\newblock {NeRF}: Representing scenes as neural radiance fields for view
  synthesis.
\newblock In {\em ECCV}, 2020.

\bibitem{miyato2018spectral}
Takeru Miyato, Toshiki Kataoka, Masanori Koyama, and Yuichi Yoshida.
\newblock Spectral normalization for generative adversarial networks.
\newblock In {\em ICLR}, 2018.

\bibitem{nakashima_deeplab}
Kazuto Nakashima.
\newblock {DeepLab with PyTorch}.
\newblock \url{https://github.com/kazuto1011/deeplab-pytorch}.

\bibitem{neff2021donerf}
Thomas Neff, Pascal Stadlbauer, Mathias Parger, Andreas Kurz, Chakravarty
  R.~Alla Chaitanya, Anton Kaplanyan, and Markus Steinberger.
\newblock {DONeRF}: Towards real-time rendering of neural radiance fields using
  depth oracle networks.
\newblock {\em arXiv preprint arXiv:2103.03231}, 2021.

\bibitem{nguyen2019hologan}
Thu Nguyen-Phuoc, Chuan Li, Lucas Theis, Christian Richardt, and Yong-Liang
  Yang.
\newblock {HoloGAN}: Unsupervised learning of 3d representations from natural
  images.
\newblock In {\em CVPR}, 2019.

\bibitem{nguyen2020blockgan}
Thu Nguyen-Phuoc, Christian Richardt, Long Mai, Yong-Liang Yang, and Niloy
  Mitra.
\newblock {BlockGAN}: Learning 3d object-aware scene representations from
  unlabelled images.
\newblock In {\em NeurIPS}, 2020.

\bibitem{niemeyer2020giraffe}
Michael Niemeyer and Andreas Geiger.
\newblock {GIRAFFE}: Representing scenes as compositional generative neural
  feature fields.
\newblock {\em arXiv preprint arXiv:2011.12100}, 2020.

\bibitem{niemeyer2020differentiable}
Michael Niemeyer, Lars Mescheder, Michael Oechsle, and Andreas Geiger.
\newblock Differentiable volumetric rendering: Learning implicit 3d
  representations without 3d supervision.
\newblock In {\em CVPR}, 2020.

\bibitem{ost2020neural}
Julian Ost, Fahim Mannan, Nils Thuerey, Julian Knodt, and Felix Heide.
\newblock Neural scene graphs for dynamic scenes.
\newblock {\em arXiv preprint arXiv:2011.10379}, 2020.

\bibitem{pan2020gan2shape}
Xingang Pan, Bo Dai, Ziwei Liu, Chen~Change Loy, and Ping Luo.
\newblock Do 2d {GANs} know 3d shape? {U}nsupervised 3d shape reconstruction
  from 2d image {GANs}.
\newblock In {\em ICLR}, 2021.

\bibitem{park2020deformable}
Keunhong Park, Utkarsh Sinha, Jonathan~T Barron, Sofien Bouaziz, Dan~B Goldman,
  Steven~M Seitz, and Ricardo-Martin Brualla.
\newblock Deformable neural radiance fields.
\newblock {\em arXiv preprint arXiv:2011.12948}, 2020.

\bibitem{park2019semantic}
Taesung Park, Ming-Yu Liu, Ting-Chun Wang, and Jun-Yan Zhu.
\newblock Semantic image synthesis with spatially-adaptive normalization.
\newblock In {\em CVPR}, 2019.

\bibitem{pumarola2020d}
Albert Pumarola, Enric Corona, Gerard Pons-Moll, and Francesc Moreno-Noguer.
\newblock {D-NeRF}: Neural radiance fields for dynamic scenes.
\newblock {\em arXiv preprint arXiv:2011.13961}, 2020.

\bibitem{rebain2020derf}
Daniel Rebain, Wei Jiang, Soroosh Yazdani, Ke Li, Kwang~Moo Yi, and Andrea
  Tagliasacchi.
\newblock {DeRF}: Decomposed radiance fields.
\newblock {\em arXiv preprint arXiv:2011.12490}, 2020.

\bibitem{riegler2020free}
Gernot Riegler and Vladlen Koltun.
\newblock Free view synthesis.
\newblock In {\em ECCV}, 2020.

\bibitem{schonfeld2021you}
Edgar Sch{\"o}nfeld, Vadim Sushko, Dan Zhang, Juergen Gall, Bernt Schiele, and
  Anna Khoreva.
\newblock You only need adversarial supervision for semantic image synthesis.
\newblock In {\em ICLR}, 2021.

\bibitem{schwarz2020graf}
Katja Schwarz, Yiyi Liao, Michael Niemeyer, and Andreas Geiger.
\newblock {GRAF}: Generative radiance fields for 3d-aware image synthesis.
\newblock In {\em NeurIPS}, 2020.

\bibitem{shu2018dirt}
Rui Shu, Hung~H Bui, Hirokazu Narui, and Stefano Ermon.
\newblock A {DIRT-T} approach to unsupervised domain adaptation.
\newblock In {\em ICLR}, 2018.

\bibitem{sitzmann2019deepvoxels}
Vincent Sitzmann, Justus Thies, Felix Heide, Matthias Nie{\ss}ner, Gordon
  Wetzstein, and Michael Zollhofer.
\newblock {DeepVoxels}: Learning persistent 3d feature embeddings.
\newblock In {\em CVPR}, 2019.

\bibitem{srinivasan2020nerv}
Pratul~P Srinivasan, Boyang Deng, Xiuming Zhang, Matthew Tancik, Ben
  Mildenhall, and Jonathan~T Barron.
\newblock {NeRV}: Neural reflectance and visibility fields for relighting and
  view synthesis.
\newblock In {\em CVPR}, 2021.

\bibitem{szegedy2016rethinking}
Christian Szegedy, Vincent Vanhoucke, Sergey Ioffe, Jon Shlens, and Zbigniew
  Wojna.
\newblock Rethinking the inception architecture for computer vision.
\newblock In {\em CVPR}, 2016.

\bibitem{tancik2020learned}
Matthew Tancik, Ben Mildenhall, Terrance Wang, Divi Schmidt, Pratul~P
  Srinivasan, Jonathan~T Barron, and Ren Ng.
\newblock Learned initializations for optimizing coordinate-based neural
  representations.
\newblock In {\em CVPR}, 2020.

\bibitem{tang2012shifting}
Kevin Tang, Vignesh Ramanathan, Li Fei-Fei, and Daphne Koller.
\newblock Shifting weights: Adapting object detectors from image to video.
\newblock In {\em NeurIPS}, 2012.

\bibitem{wang2018high}
Ting-Chun Wang, Ming-Yu Liu, Jun-Yan Zhu, Andrew Tao, Jan Kautz, and Bryan
  Catanzaro.
\newblock High-resolution image synthesis and semantic manipulation with
  conditional {GANs}.
\newblock In {\em CVPR}, 2018.

\bibitem{wiles2019synsin}
Olivia Wiles, Georgia Gkioxari, Richard Szeliski, and Justin Johnson.
\newblock {SynSin}: End-to-end view synthesis from a single image.
\newblock In {\em CVPR}, 2020.

\bibitem{xian2020space}
Wenqi Xian, Jia-Bin Huang, Johannes Kopf, and Changil Kim.
\newblock Space-time neural irradiance fields for free-viewpoint video.
\newblock {\em arXiv preprint arXiv:2011.12950}, 2020.

\bibitem{xie2018learning}
Shaoan Xie, Zibin Zheng, Liang Chen, and Chuan Chen.
\newblock Learning semantic representations for unsupervised domain adaptation.
\newblock In {\em ICML}, 2018.

\bibitem{yariv2020multiview}
Lior Yariv, Yoni Kasten, Dror Moran, Meirav Galun, Matan Atzmon, Basri Ronen,
  and Yaron Lipman.
\newblock Multiview neural surface reconstruction by disentangling geometry and
  appearance.
\newblock In {\em NeurIPS}, 2020.

\bibitem{yarowsky1995unsupervised}
David Yarowsky.
\newblock Unsupervised word sense disambiguation rivaling supervised methods.
\newblock In {\em ACL}, 1995.

\bibitem{yuan2020star}
Wentao Yuan, Zhaoyang Lv, Tanner Schmidt, and Steven Lovegrove.
\newblock {STaR}: Self-supervised tracking and reconstruction of rigid objects
  in motion with neural rendering.
\newblock {\em arXiv preprint arXiv:2101.01602}, 2020.

\bibitem{zhang2020nerf}
Kai Zhang, Gernot Riegler, Noah Snavely, and Vladlen Koltun.
\newblock {NeRF++}: Analyzing and improving neural radiance fields.
\newblock {\em arXiv preprint arXiv:2010.07492}, 2020.

\bibitem{zhang2018collaborative}
Weichen Zhang, Wanli Ouyang, Wen Li, and Dong Xu.
\newblock Collaborative and adversarial network for unsupervised domain
  adaptation.
\newblock In {\em CVPR}, 2018.

\bibitem{zhang2021image}
Yuxuan Zhang, Wenzheng Chen, Huan Ling, Jun Gao, Yinan Zhang, Antonio Torralba,
  and Sanja Fidler.
\newblock Image {GAN}s meet differentiable rendering for inverse graphics and
  interpretable 3d neural rendering.
\newblock In {\em ICLR}, 2021.

\bibitem{zhu2017unpaired}
Jun-Yan Zhu, Taesung Park, Phillip Isola, and Alexei~A Efros.
\newblock Unpaired image-to-image translation using cycle-consistent
  adversarial networks.
\newblock In {\em ICCV}, 2017.

\bibitem{zhu2017toward}
Jun-Yan Zhu, Richard Zhang, Deepak Pathak, Trevor Darrell, Alexei~A Efros,
  Oliver Wang, and Eli Shechtman.
\newblock Toward multimodal image-to-image translation.
\newblock In {\em NeurIPS}, 2017.

\bibitem{zou2018unsupervised}
Yang Zou, Zhiding Yu, BVK Vijaya~Kumar, and Jinsong Wang.
\newblock Unsupervised domain adaptation for semantic segmentation via
  class-balanced self-training.
\newblock In {\em ECCV}, 2018.

\end{thebibliography}
}

\twocolumn[{%
\renewcommand\twocolumn[1][]{#1}%
\begin{center}
    {\bf \Large GANcraft: Unsupervised 3D Neural Rendering of Minecraft Worlds\\[4pt]
    Supplementary Material}
\end{center}
}]

\setcounter{section}{0}
\renewcommand\thesection{\Alph{section}}

\section{Supplementary video}
Our project website is available at \href{https://nvlabs.github.io/GANcraft/}{\texttt{https://nvlabs.github.io/GANcraft/}}. This includes an overview of the method as well as additional results.

We also provide a video, including more visual results and discussion of our work. Specifically, it contains:
\setitemize{noitemsep,topsep=2pt,parsep=2pt,partopsep=0pt}
\begin{itemize}
    \item Additional high-resolution video results rendered at 1024$\times$2048 pixels and 30 frames per second
    \item Style interpolation results
    \item Additional comparisons with baseline methods
    \item Illustration of the proposed approach.
\end{itemize}
Please make sure to check it out at\\
\href{https://www.youtube.com/watch?v=1Hky092CGFQ}{https://www.youtube.com/watch?v=1Hky092CGFQ}.

\section{Method details}

Here, we provide additional details of our approach.
\subsection{Numerical volumetric rendering}
The integral in Equation~2 of the main paper can be approximated with discrete samples via quadrature \cite{mildenhall2020nerf}.
Assume that we sample $N+1$ points at $t_1, ..., t_{N+1}$ along a camera ray $\br(t) = \bo + t \bv$. We define
\begin{gather*}
    \delta_i = t_{i+1} - t_i, \\
    \hat{t}_i = \frac{t_{i+1} + t_i}{2}, \\
    \sigma_i = \sigma\left(\br\left(\hat{t}_i\right), l\left(\br\left(\hat{t}_i\right)\right)\right), \\
    \bc_i = \bc\left(\br\left(\hat{t}_i\right), l\left(\br\left(\hat{t}_i\right)\right), \bz \right),
\end{gather*}
such that
\begin{gather*}
    C(\br) \approx \left\{ \sum_{i=1}^{N} T_i (1 - \exp(-\sigma_i \delta_i)) \bc_i \right\} + T_{N+1} \bc_\mathrm{sky}(\bv, \bz), \\
    \mathrm{where}~ T_i = \exp{\left( - \sum_{j=1}^{i-1} \sigma_j \delta_j \right)}.
\end{gather*}

\subsection{Point sampling algorithm}
In this section, we describe the method we use to efficiently sample points from the sparse voxel grid along a camera ray. Instead of relying on rejection sampling (as in Liu~\etal~\cite{liu2020neuralsparse}) to remove points that have not landed inside any voxel, we first traverse the voxel grid along the ray to obtain the entrance and exit points of each valid voxel that the ray has gone through, and then sample points only on the segments that are inside voxels.

For voxel grid traversal, We implement a 3D version of Bresenham's line algorithm~\cite{amanatides1987fast}, which has a very low computational cost of $O(N)$, where $N$ is the longest dimension of the voxel grid. Its working principle is as follows: Starting from the voxel position where the camera resides, for each step, we traverse to the next voxel which is adjacent to the current voxel by the face which the ray exits from.

\subsection{Network Architecture}
\begin{figure*}[h!]
    \centering
    \includegraphics[width=0.86\textwidth,trim={0.1in 3.4in 1.45in 0.0in},clip]{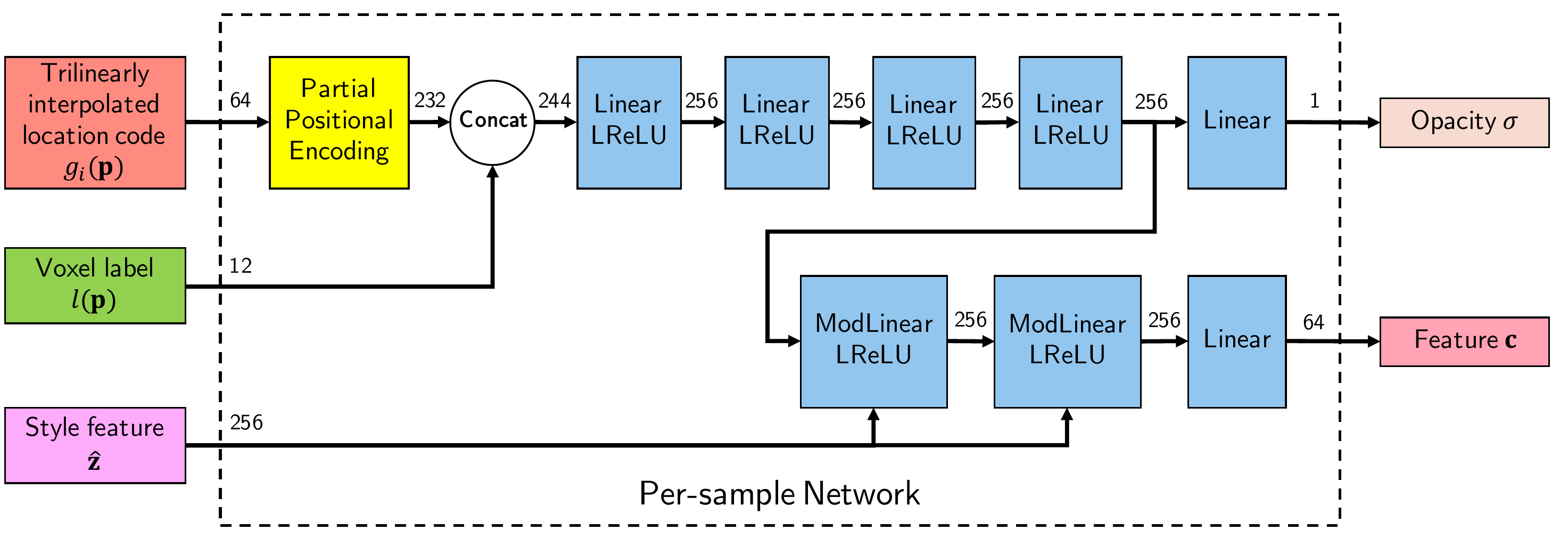}
    \caption{Per-sample MLP for representing the implicit radiance field in conjunction with the voxel features. We use weight modulation to condition the output feature $\bc$ on the style feature. This is more computationally efficient than doing affine modulation on the per-layer feature when the same style is applied to a large number of samples. The number on each arrow denotes the number of channels. As a means of conserving the memory, we use partial positional encoding on the location code, which performs positional encoding only on the first 24 channels, and concatenate the result with the remaining 40 channels.}
    \label{fig:3d_point_encoder}

    \vspace{20pt}

    \includegraphics[width=0.91\textwidth,trim={0.1in 5.0in 0.7in 0.1in},clip]{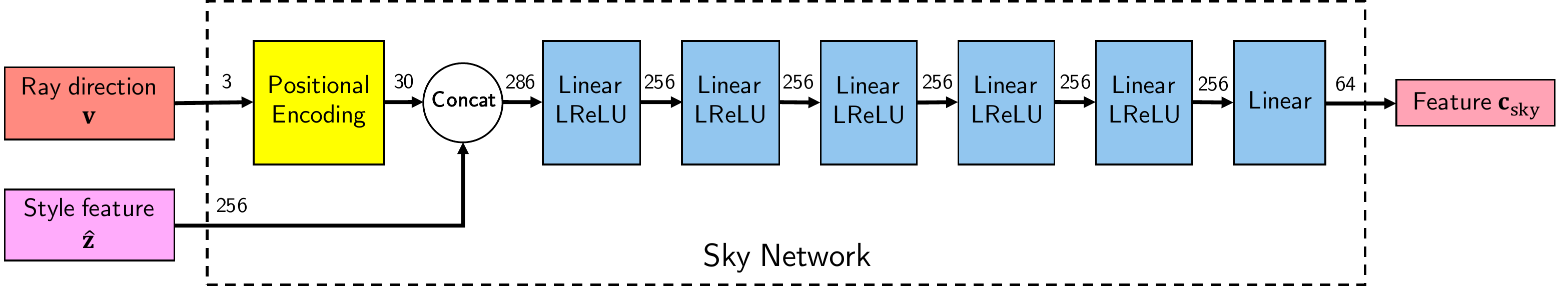}
    \caption{Network architecture for the neural sky dome MLP. The input ray direction is represented as a normalized 3D vector. The numbers on the arrows denote the number of channels.}
    \label{fig:sky_mlp}

    \vspace{25pt}

    \includegraphics[width=0.9\textwidth,trim={0.15in 4.6in 1.0in 0.0in},clip]{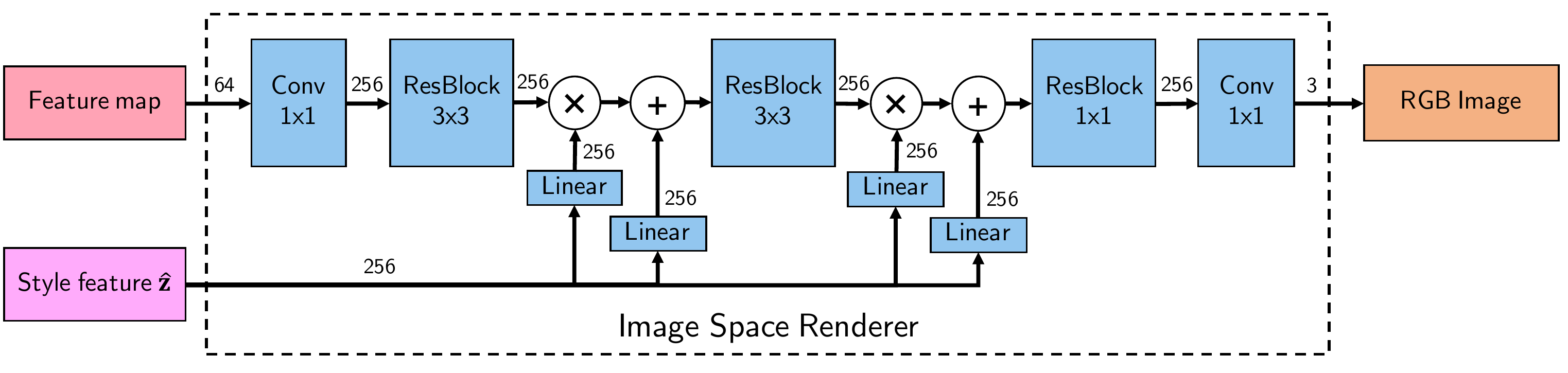}
    \caption{Network architecture for the image space renderer. The kernel sizes are shown inside each block and the channel counts are displayed on the arrows. We apply Leaky ReLU after each Conv block, inside ResBlocks and after the affine modulations. We use hyperbolic tangent activation for generating the final image (omitted here for clarity).}
    \label{fig:image_generator}

    \vspace{20pt}

    \includegraphics[width=0.72\textwidth,trim={0.1in 5.2in 3.9in 0.1in},clip]{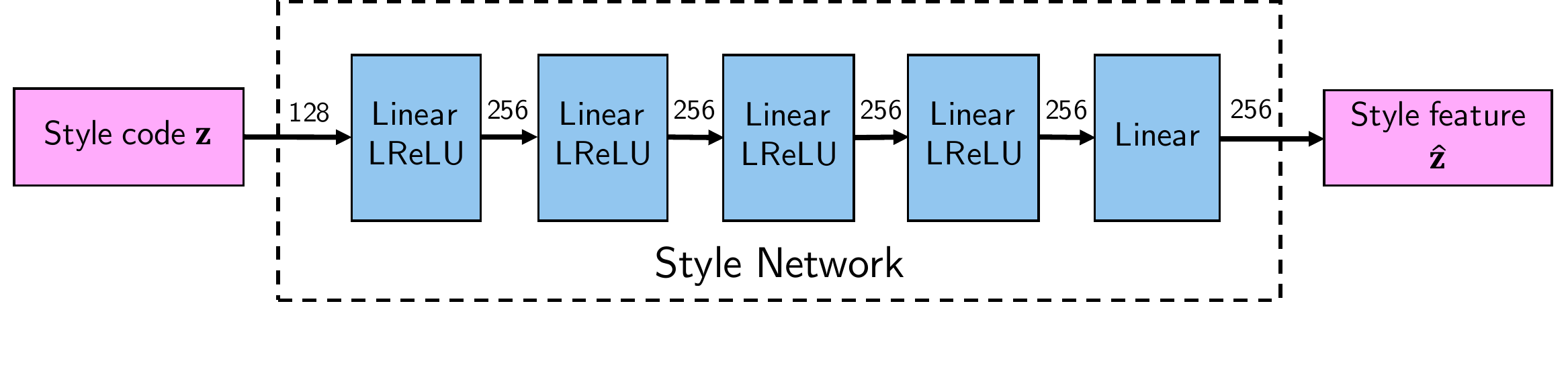}
    \caption{The architecture of style network. Following StyleGAN2~\cite{karras2020analyzing}, we use a common MLP that is shared across all the style conditioning layers to convert the input style code to an intermediate style feature.}
    \label{fig:style_net}
\end{figure*}

\begin{figure*}[htb!]
    \centering
    \includegraphics[width=0.75\textwidth,trim={0.1in 3.4in 3.2in 0.1in},clip]{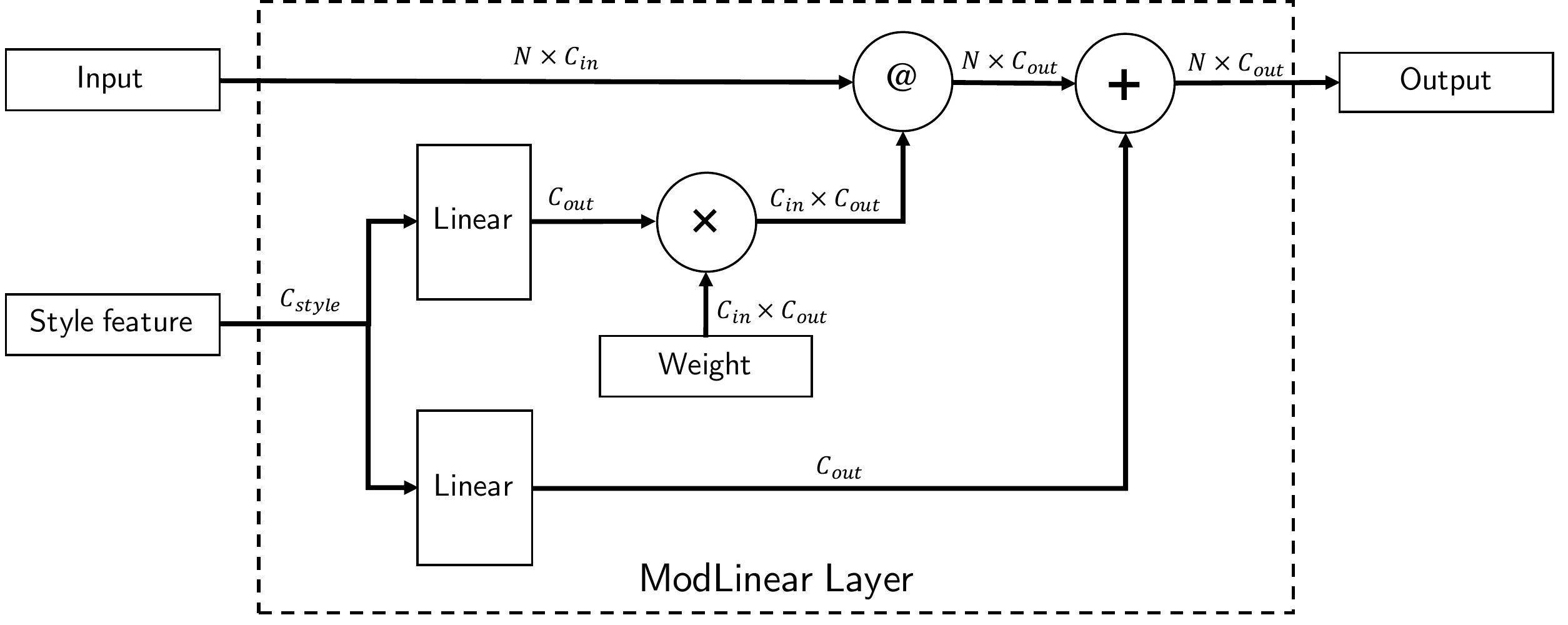}
    \caption{Detailed structure of ModLinear layer used in the per-sample network (Figure~\ref{fig:3d_point_encoder}). `@' denotes matrix multiplication. Shapes of intermediate tensors are denoted on the arrows. The batch dimension is omitted for clarity.}
    \label{fig:modlinear}
\end{figure*}

GANcraft contains 6 trainable neural networks. Here are their descriptions and their respective network architectures:

\medskip
\noindent{\bf Per-sample MLP.} This is the MLP for representing the implicit radiance field, in conjunction with the voxel features. The network architecture is illustrated in Fig.~\ref{fig:3d_point_encoder}. We condition the output feature on the style code via weight modulation \cite{karras2020analyzing}. The detailed implementation of weight modulation is shown in Fig.~\ref{fig:modlinear}.

\medskip
\noindent{\bf Neural sky dome.} The sky is modeled with an MLP (Fig.~\ref{fig:sky_mlp}) which takes ray direction (represented as a normalized 3D vector) input and produce the color feature for that ray. The network is also conditional on the style feature.

\medskip
\noindent{\bf Image space renderer.} This is a CNN for converting feature map to RGB image (Fig.~\ref{fig:image_generator}). As discussed in the main paper, we use very small kernel sizes to reduce the receptive field in order to encourage view consistency. The network is conditional on the style feature.

\medskip
\noindent{\bf Style network.} Following StyleGAN2~\cite{karras2020analyzing}, we use an MLP that is shared across all the style conditioning layers to convert the input style code to an intermediate style feature. Its architecture is shown in Fig.~\ref{fig:style_net}.

\medskip
\noindent{\bf Style encoder.} The style encoder is a CNN that predicts the style code given an image. In conjunction with pseudo ground truth and reconstruction loss, this allows GANcraft to produce images that follows the style of a given image. Our style encoder is taken from SPADE~\cite{park2019semantic}, which is a 6-layer CNN followed by a linear layer and VAE reparameterization. Please refer to the original paper for the details.

\begin{figure*}[htb!]
    \centering
    \includegraphics[width=0.99\textwidth,trim={0.0in 1.7in 0.0in 0.0in},clip]{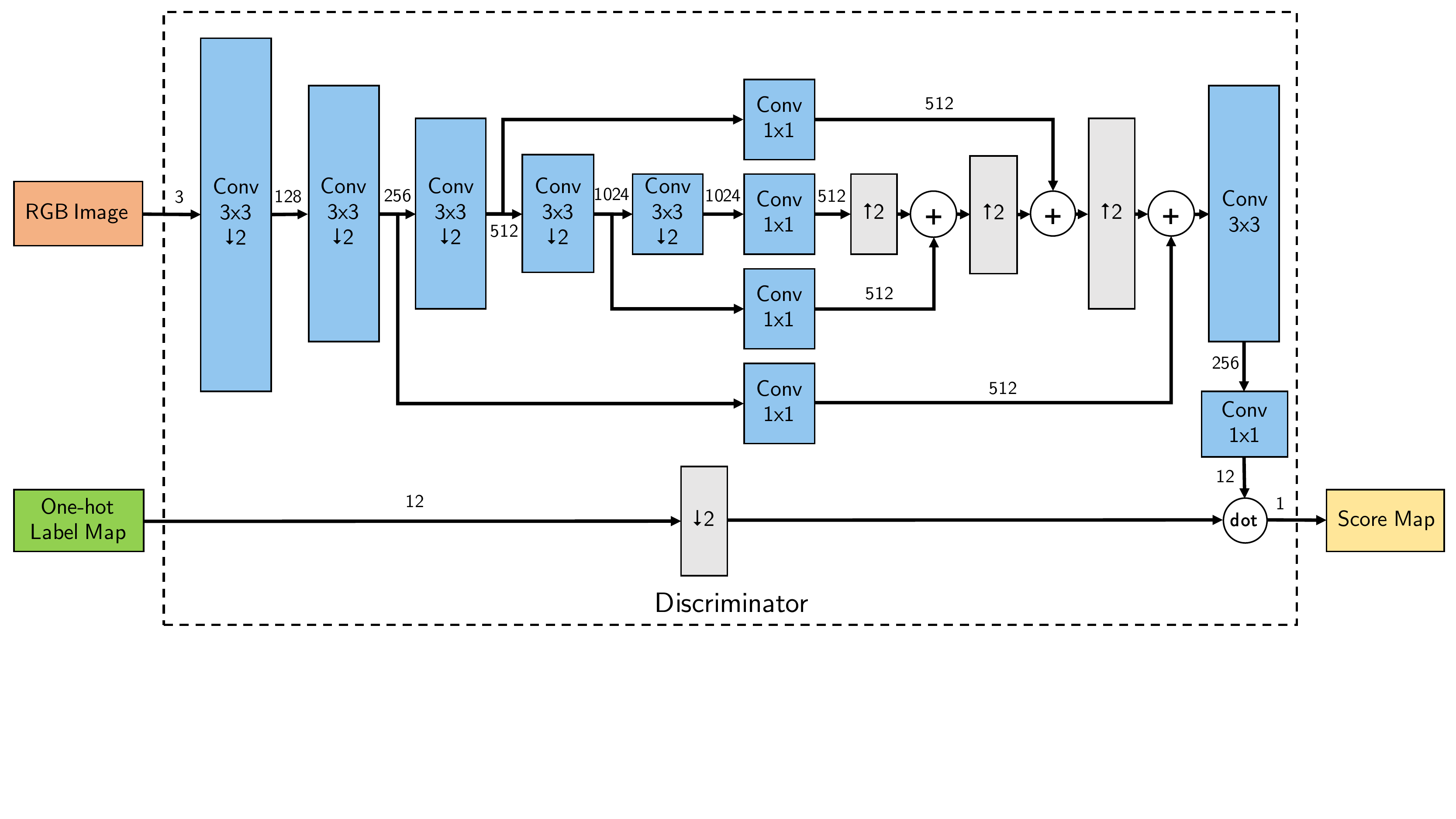}
    \caption{The conditional discriminator used in GANcraft. `dot' denotes dot product on the channel dimension. `$\downarrow2$' denotes downsample by 2. `$\uparrow2$' denotes upsample by 2. We use bilinear interpolation for upsampling, and stride 2 convolution for downsampling. For label map, we downsample it via nearest neighbor interpolation. We use spectral normalization~\cite{miyato2018spectral} on all the convolution layers in the discriminator.}
    \label{fig:discriminator}
\end{figure*}

\medskip
\noindent{\bf Label-conditional discriminator.} The discriminator we use is based on feature pyramid semantics-embedding (FPSE) discriminator~\cite{liu2019learning}. Its construction is shown in Fig.~\ref{fig:discriminator}. Compared to the patch discriminator used in SPADE~\cite{park2019semantic}, the FPSE discriminator is more robust to the distribution mismatch in the label map domain. A patch discriminator which takes the concatenated image and label map as input sometimes lead to training collapse almost immediately after the training starts.

\begin{figure*}
    \centering
    \setlength{\tabcolsep}{0pt}
    \begin{adjustbox}{max width=\textwidth}
    \begin{tabular}{ccccc}
         \includegraphics[width=0.2\textwidth]{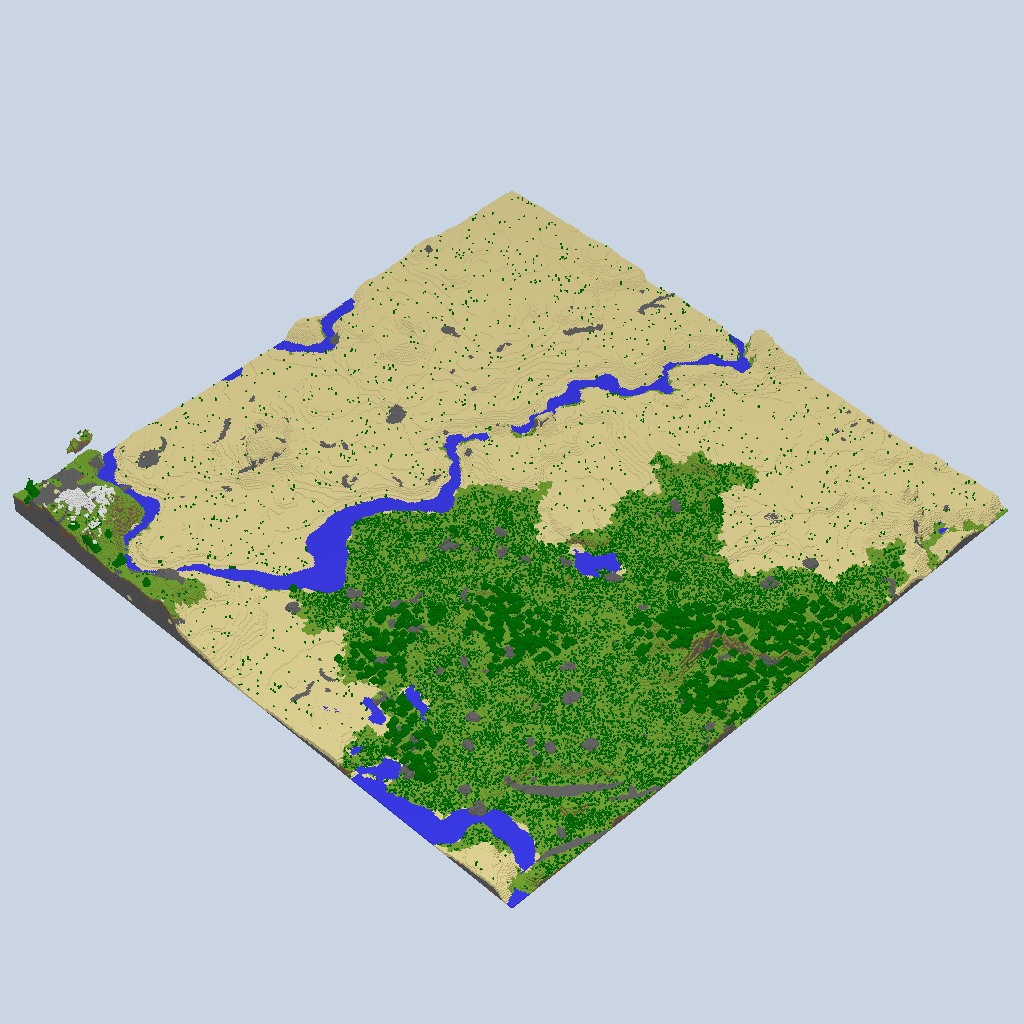} &
         \includegraphics[width=0.2\textwidth]{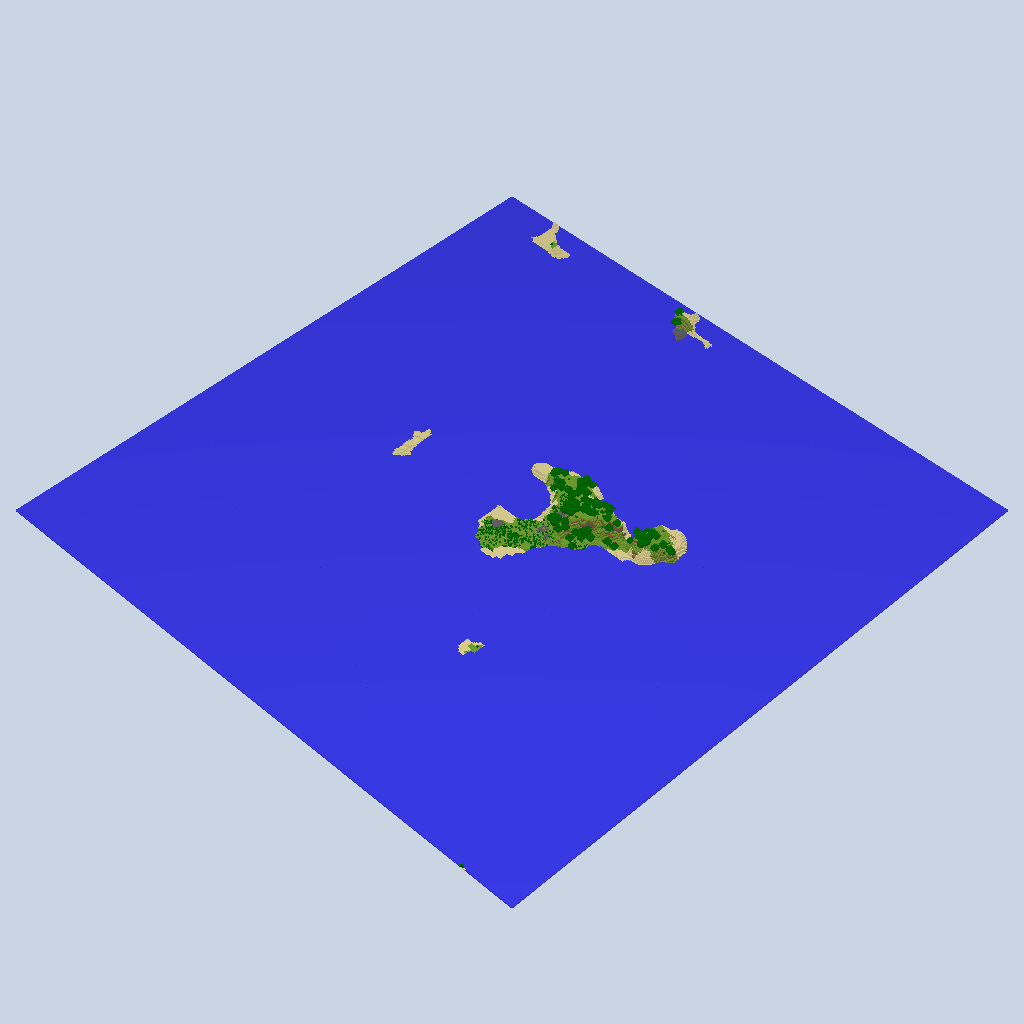} &
         \includegraphics[width=0.2\textwidth]{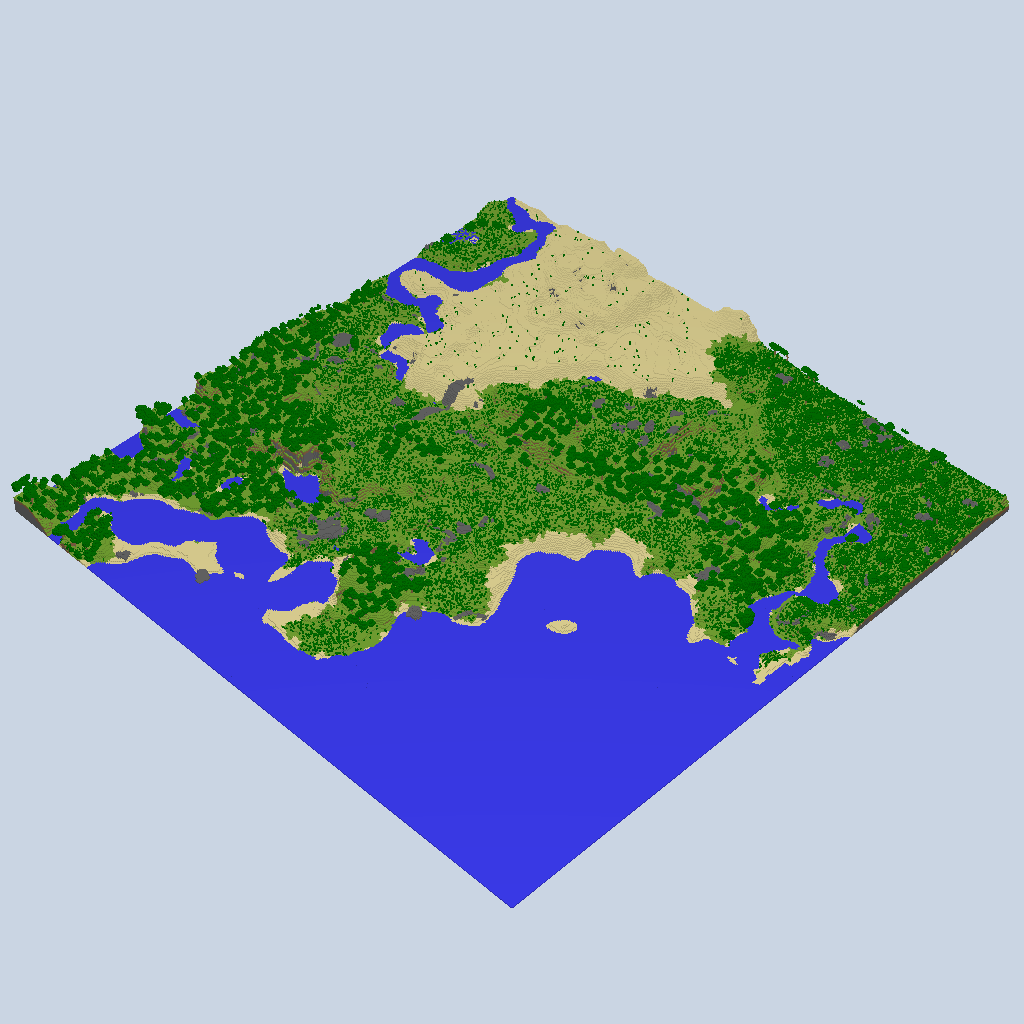} &
         \includegraphics[width=0.2\textwidth]{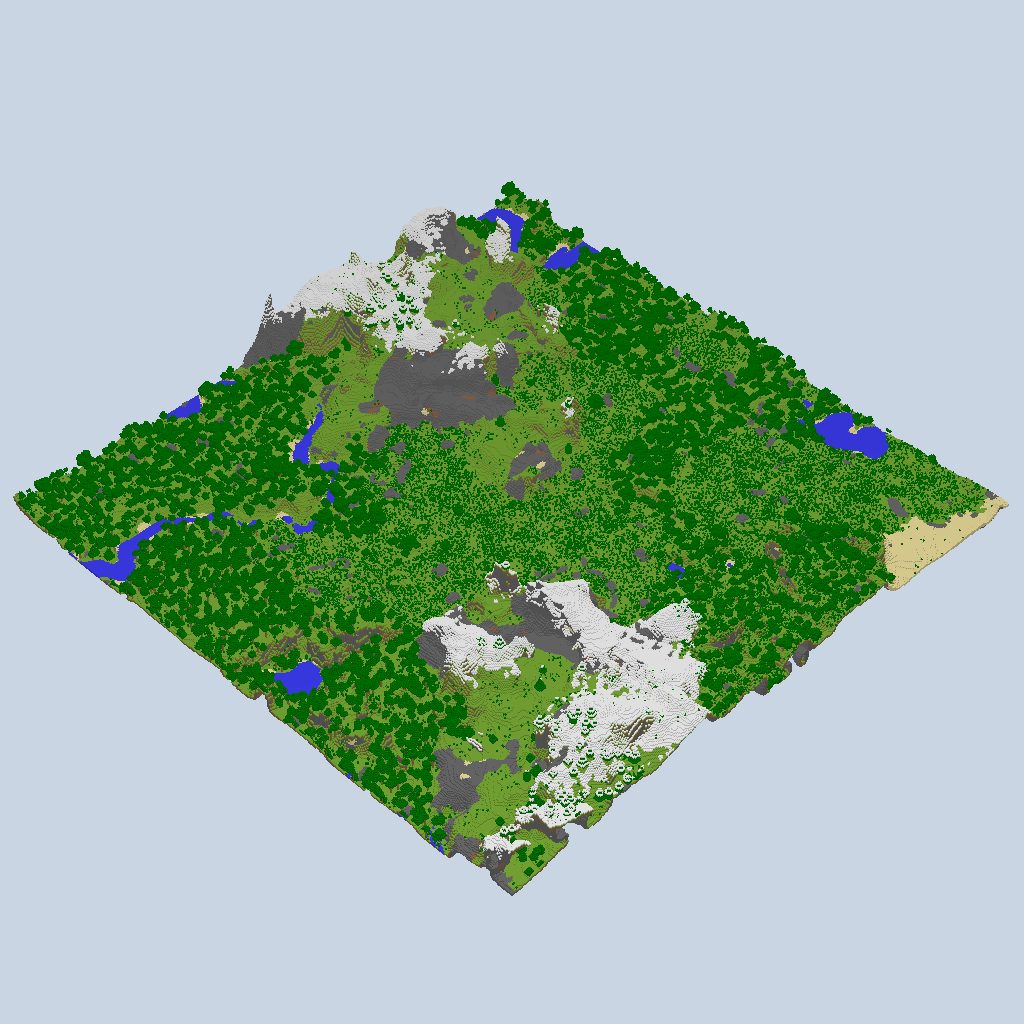} &
         \includegraphics[width=0.2\textwidth]{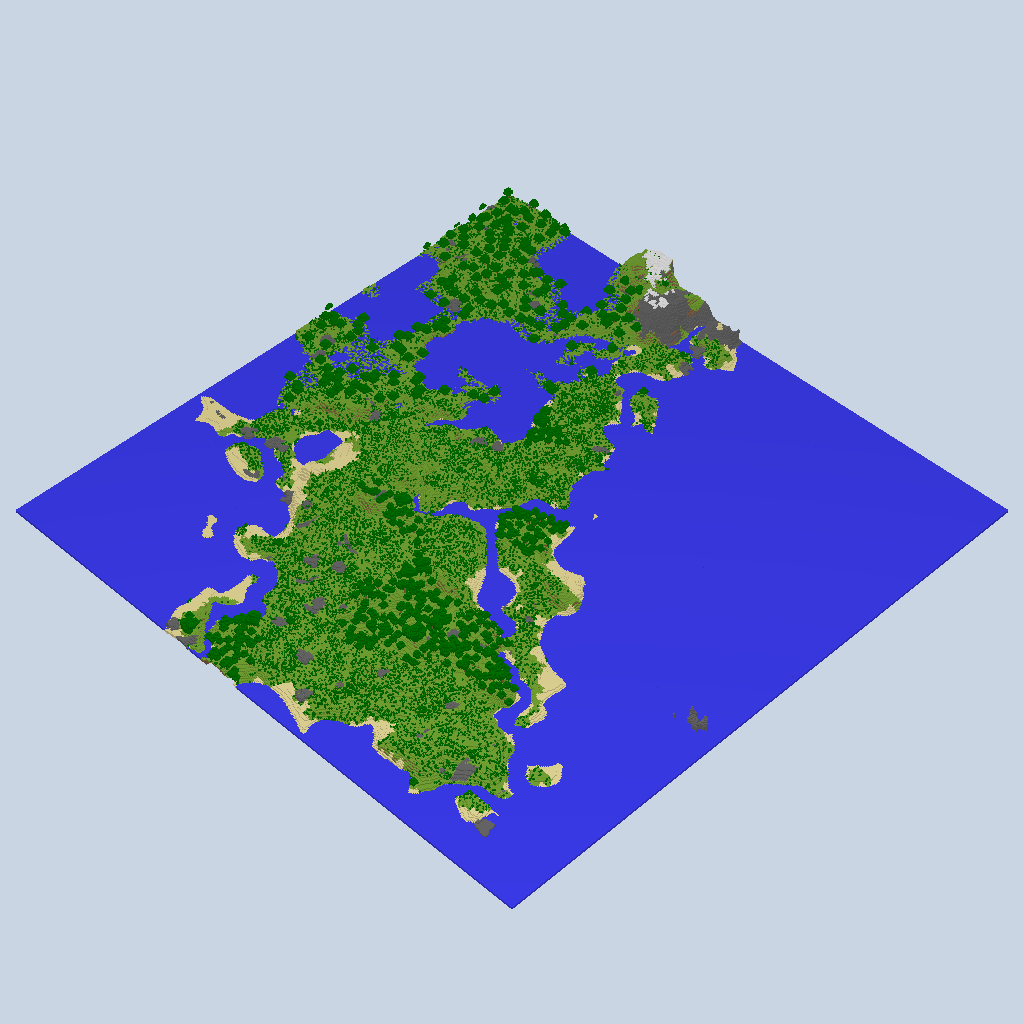}
    \end{tabular}
    \end{adjustbox}
    \caption{{\bf Bird's-eye view of the 5 Minecraft worlds used.} Each block is color-coded by its label (brown-sand, blue-water, light green-grass, dark green-trees, white-snow, \etc). We use worlds with varying distributions of sand, forest, water, snow, trees, grass, \etc.
    The label distribution of each specific world is very different from that of a collection of real images, \eg the first world is $>$50\% sand, and the second is $>$90\% water. Our method works for all these worlds despite the domain gap, indicating the robustness of our framework.}
    \label{fig:worlds}
\end{figure*}

\begin{figure*}[h!]
    \centering
    \begin{adjustbox}{max width=\textwidth}
        \includegraphics[width=\textwidth]{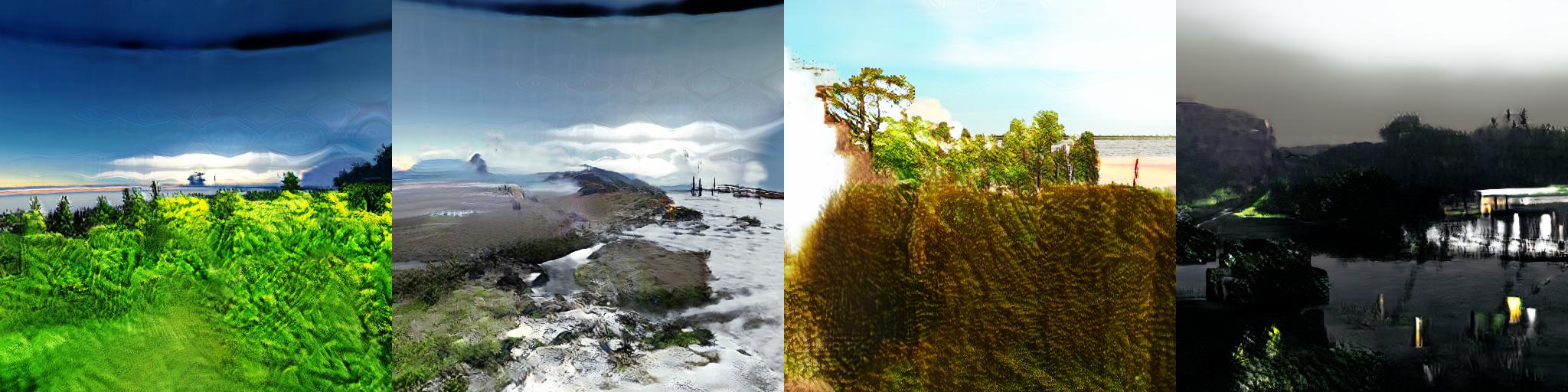}
    \end{adjustbox}
    \caption{{\bf Outputs of the ablated model that does not use pseudo-ground truths.}
    This model was trained only with a GAN loss between the outputs and real images, and obtains low FID and KID values, as seen in Table~\ref{table:ablation_metrics}. However, the output images look unrealistic and do not learn the correct correspondence between input segmentation labels and realistic textures.}
    \label{fig:no_pseudo_gt}
    \vspace{5mm}
    \centering
    \setlength{\tabcolsep}{0pt}
    \begin{adjustbox}{max width=\textwidth}
    \begin{tabular}{cccc}
         \includegraphics[width=0.25\textwidth]{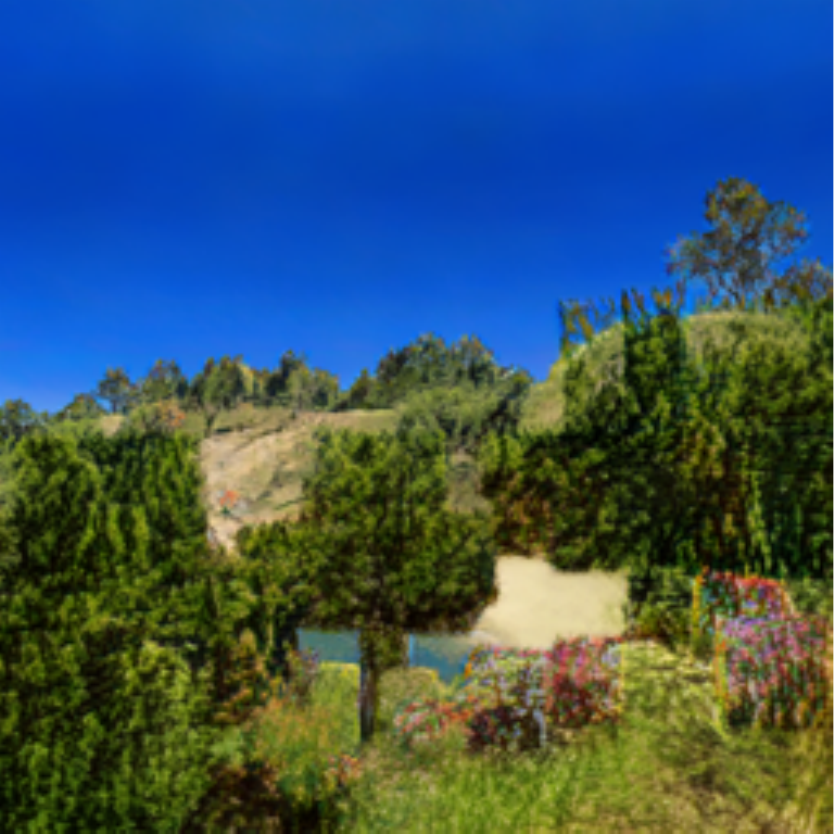} &
         \includegraphics[width=0.25\textwidth]{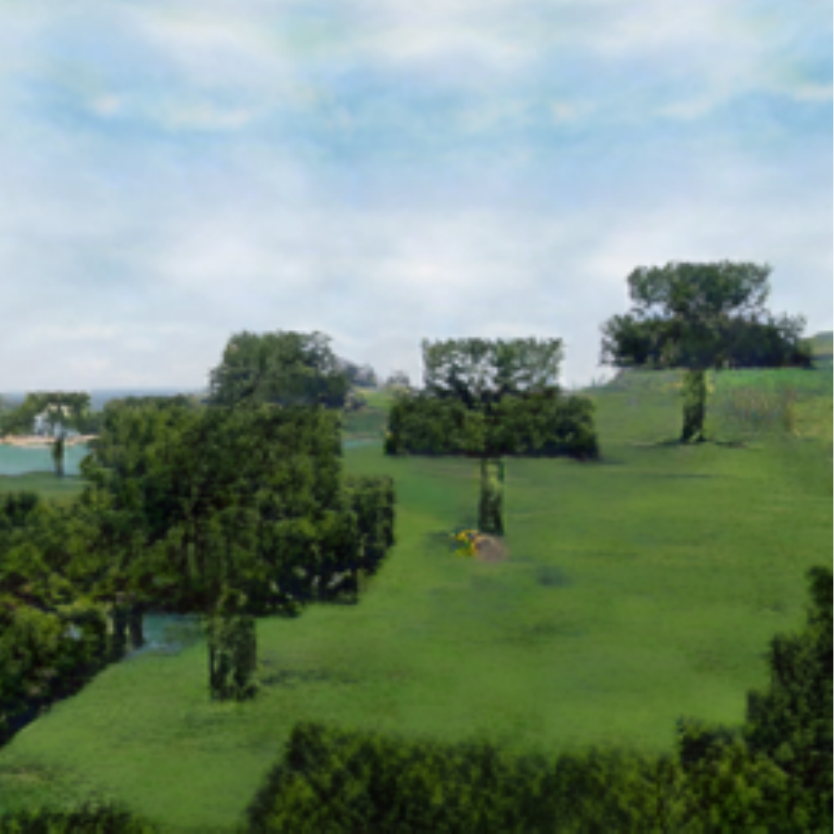} &
         \includegraphics[width=0.25\textwidth]{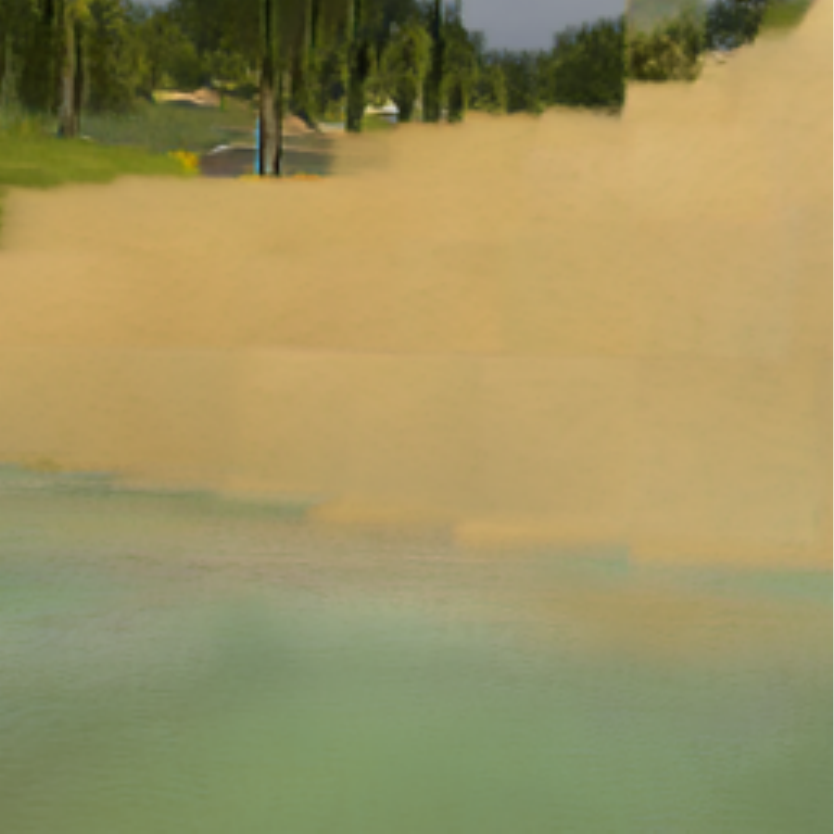} &
         \includegraphics[width=0.25\textwidth]{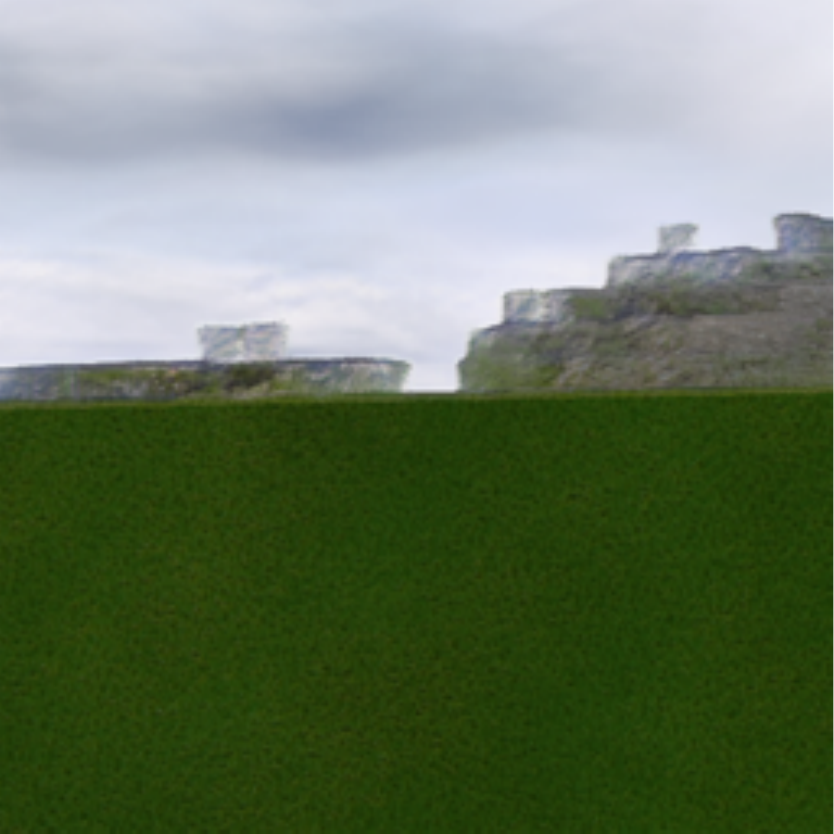}
    \end{tabular}
    \end{adjustbox}
    \caption{{\bf Blockiness in some outputs.} Certain regions and objects appear blocky due to the underlying blocky geometry that is very different from occurrences in the real world.}
    \label{fig:failure_blockiness}
\end{figure*}

\begin{figure}[h!]
    \centering
    \setlength{\tabcolsep}{1pt}
    \begin{adjustbox}{max width=\columnwidth}
    \begin{tabular}{cccc}
         \includegraphics[width=0.5\columnwidth]{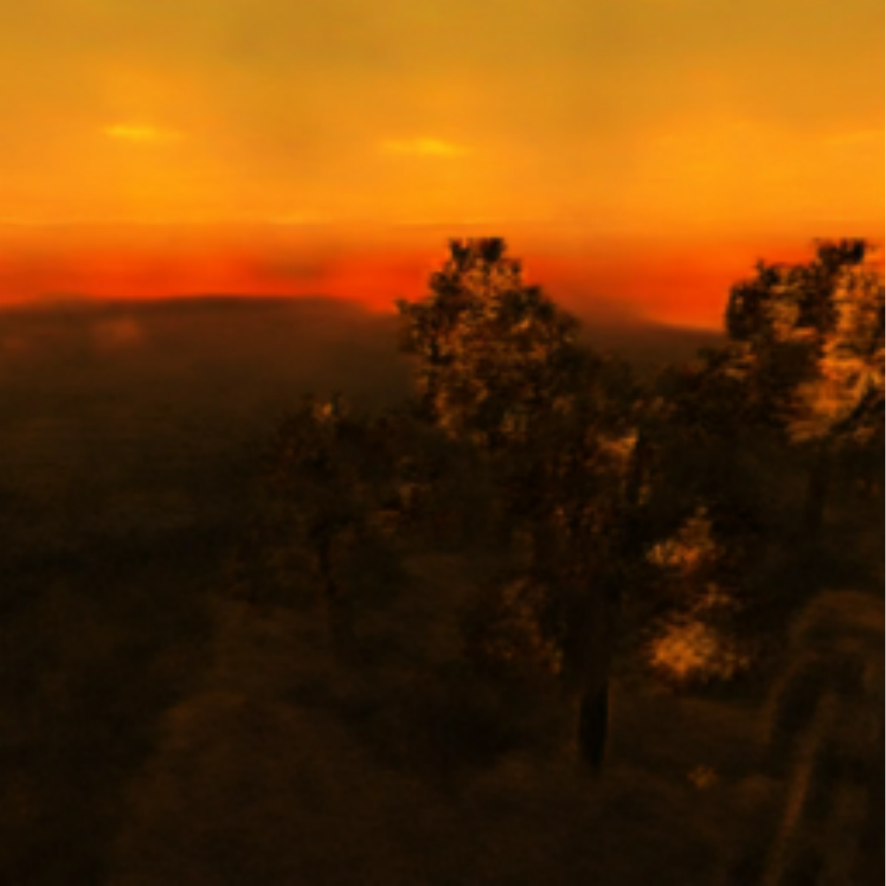} &
         \includegraphics[width=0.5\columnwidth]{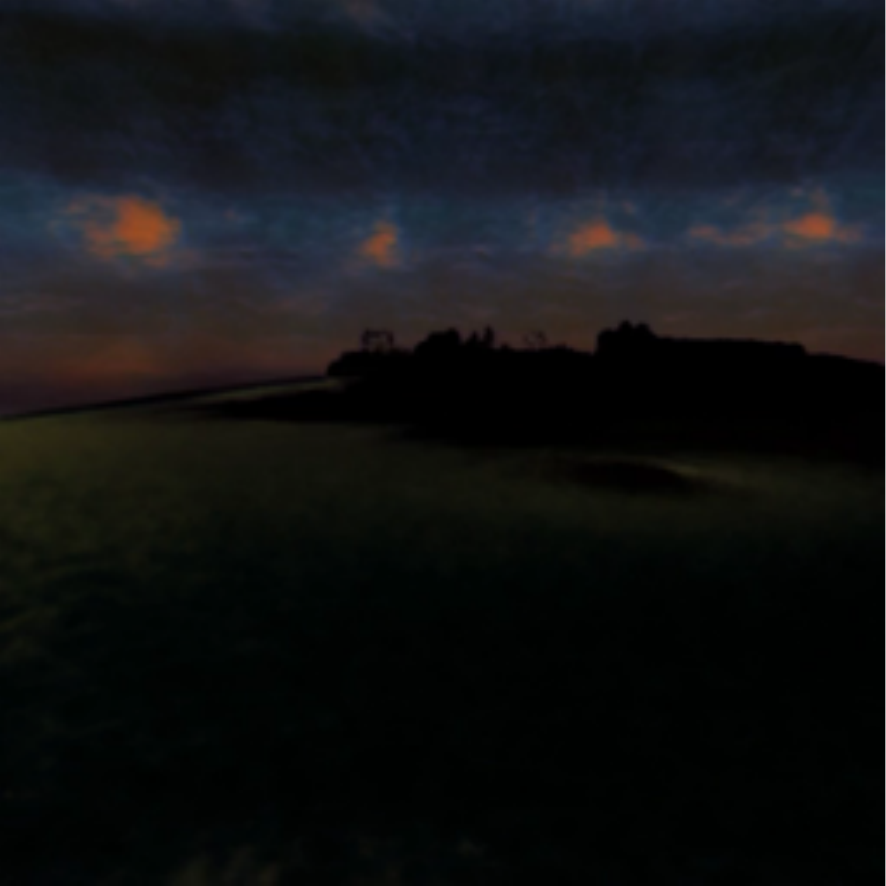}
    \end{tabular}
    \end{adjustbox}
    \caption{{\bf Incompatible styles.} Certain combinations of styles and worlds give unrealistic outputs, possibly as these styles are outliers.}
    \label{fig:failure_style}
\end{figure}

\subsection{Label Translation}
There is significant difference between Minecraft voxel labels, which we use as the starting point of GANcraft, and COCO-Stuff~\cite{caesar2018coco} labels, which is the format the pretrained DeepLabV2 model produces and the pretrained SPADE model accepts.
In Minecraft Java edition, there are 680 labels in total, mostly describing raw materials (dirt, sand, log, water, etc.) useful for building objects. While in COCO-Stuff, there are 182 higher level labels of common objects such as mountain, tree, river, and sea.
Due to the drastic difference in the level of abstraction, it is very difficult to find a one-to-one mapping between Minecraft label and COCO-Stuff label. For example, the water material in Minecraft can be mapped to either sea or river label in COCO-Stuff; the tree material consists of both log and leaf label in COCO-Stuff.
We solve the labeling difference in two ways. For the label-conditional discriminator, we introduce a new set of 12 classes with high level of abstraction: \textit{ignore, sky, tree, dirt, flower, grass, gravel, water, rock, stone, sand, and snow}. We then classify every Minecraft and COCO-Stuff label into one of the 12 classes, and use the translated semantic segmentation mask as the conditional input to the discriminator.
For generating pseudo-ground truth, however, we will have to convert Minecraft labels to COCO-Stuff labels in order to be recognized by the pretrained SPADE generator. We achieve this by first translating the Minecraft labels to one of the 12 labels, and then map them to COCO-Stuff labels randomly, with equal chance across all the candidate labels. Note that we use the same mapping scheme within a segmentation map. We are able to obtain good result from such a simple measure, as the style encoder is able to explain away the randomness in the mapping.

\subsection{Voxel Preprocessing}
Minecraft voxel world has a sea level of 62, below which most of the voxels are not visible from above. It will be a waste of memory if we still assign voxel features to those invisible voxels. Thus we preprocess the voxel by removing the interior voxels, leaving a 4 voxel thick thin shell. This operation reduces the occupancy of a typical voxel world from 28\% to 3\%. The effect of preprocessing can be seen at the borders of the voxel worlds in Fig.~\ref{fig:worlds}. Note that the preprocessing step is not only useful for Minecraft world. It is applicable to any types of voxel grids.

\section{Experiment details}

\subsection{Minecraft block worlds.}

We use 5 different Minecraft worlds for our experiments. An overview of these worlds is shown in Fig.~\ref{fig:worlds}. As can be seen, the label distribution of each specific world is very different from that of a collection of real images, \eg the first world is $>$50\% sand, and the second is $>$90\% water. Our method works for all these worlds despite the domain gap, indicating the robustness of our framework.

\subsection{GANcraft settings}
During training, we generate images at a resolution of 256$\times$256. We sample 24 points along each ray, and truncate the rays to a maximum distance of 3 (distance traveled outside voxels doesn't count). We use a learning rate of 1e-4 for the generator networks, and 4e-4 for the discriminator. For voxel features, we use a higher learning rate of 5e-3. We use a combination of GAN loss, L$_2$ loss, L$_1$ loss and perceptual loss, with their weights being 1.0, 10.0, 1.0 and 10.0, respectively. For regularization terms, we use a weight of 0.5 for the opacity regularization, and a weight of 0.05 for the KL divergence needed by the style encoder. We also clip the per-sample feature $\bc$ to a range of $[-1, 1]$ before blending to reduce the ambiguity between the opacity and the scale of feature. For random camera pose sampling, we sample two points that are slightly above ground, and use one of the as the camera location and the other one as the point that the camera looks at. We reject any camera pose that produces a depth map with a mean depth below 2 or that produces a segmentation mask with label entropy below 0.75. This guarantees that the segmentation mask along can provide enough scene geometry hint to the SPADE generator for generating a pseudo-ground truth that corresponds well to the actual scene geometry.

During evaluation, we increase the sample count to 32 points per ray. On an NVIDIA Titan V, this takes approximately 10 seconds to render a 1024$\times$2048 frame.

\subsection{Baseline settings}
For fair comparison, the settings used in the NSVF-W baseline largely resembles GANcraft except for the following differences:
\begin{itemize}
    \item Only L$_2$ loss and KL divergence is used during training.
    \item The weight for KL divergence is reduced to 0.01 to avoid handicapping the style encoder too much in the absence of other reconstruction losses.
    \item The image space CNN renderer is removed, and the per-sample MLP directly produces an RGB radiance (clipped by a sigmoid function) instead of a feature.
\end{itemize}

\section{Additional results}

\begin{table}[h!]
    \centering
    \begin{adjustbox}{max width=\textwidth}
    \begin{tabular}{lcc}
        \toprule
        Method & FID $\downarrow$ & KID $\downarrow$ \\
        \midrule
        Full model & 78.79 & 0.043 \\
        No CNN & 84.86 & 0.049 \\
        No real images & 89.95 & 0.055 \\
        No GAN loss & 104.58 & 0.073 \\
        No pseudo-ground truth & 65.40 & 0.043 \\
        \bottomrule
    \end{tabular}
    \end{adjustbox}
    \caption{{\bf Ablation comparison on automated image quality metrics} ($\downarrow$ indicates lower is better){\bf .} We compare ablated versions of our full method on a single block world.}
    \label{table:ablation_metrics}
\end{table}

\subsection{Ablation study}
Here, we present quantitative results for the ablated versions of our full model. Sample outputs from these ablations were shown in Fig. 5 of the main paper. We trained all ablations on one world only, due to computational constraints (each model takes 4 days on 8 NVIDIA V100 GPUs).

The results of automated metric evaluation as shown in Table~\ref{table:ablation_metrics}. We computed the FID and KID values with 2000 images generated from random camera poses and 5000 held-out real images.
As expected, all ablated versions obtain higher FID and KID scores indicating worse quality. An exception is the model trained without any pseudo-ground truth images, \ie trained with GAN loss between outputs and real images only. Surprisingly, it obtains a lower FID and KID than our full model. However, when we visually inspect the outputs, shown in Fig.~\ref{fig:no_pseudo_gt}, it is clear that the model fails to learn a meaningful mapping from Minecraft segmentations to real images. The model seems to have learned to produce unrealistic images that optimize the metrics due to training with the GAN loss. However, similar to MUNIT~\cite{huang2018multimodal}, the outputs are both unrealistic and incorrectly map Minecraft segmentation labels to real images.

We observed that our method can fail in two ways --- either producing blocky outputs or producing unrealistic outputs. In the input block world, all objects and regions are made of blocks. Due to this coarse geometry, the method is sometimes unable to learn realistic geometries in the translated world. As a result, boundaries can often appear jagged, as shown in Fig.~\ref{fig:failure_blockiness}. Further, certain combinations of worlds and style-conditioning images can produce unrealistic outputs as shown in Fig.~\ref{fig:failure_style}. For example, a forest world paired with a conditioning image of a red sunset can produce unrealistic, or overly dark outputs. As the style encoder is trained exclusively with pseudo-ground truth images that have the same label distribution as the rendered Minecraft images, it has never encountered such combinations.

\end{document}